%% file: main_camera_ready.tex
\documentclass[10pt,twocolumn,letterpaper]{article}

\usepackage{cvpr}              

\input{preamble}

\definecolor{cvprblue}{rgb}{0.21,0.49,0.74}
\usepackage[pagebackref,breaklinks,colorlinks,citecolor=cvprblue]{hyperref}

\def\paperID{140} 
\def\confName{3DV\xspace}
\def\confYear{2025\xspace}

\newcommand\blfootnote[1]{%
  \begingroup
  \renewcommand\thefootnote{}\footnote{#1}%
  \addtocounter{footnote}{-1}%
  \endgroup
}

\input{sec/0_macros}

\begin{document}

\title{PIR: Photometric Inverse Rendering with Shading Cues Modeling\\and Surface Reflectance Regularization}
\vspace{-3mm}
\author{
Jingzhi Bao$^{1,2}$ \hspace{0.5cm} 
Guanying Chen$^{3}$\thanks{Corresponding author} \hspace{0.5cm}
Shuguang Cui$^{4,1}$
\\
$^1$ FNii-Shenzhen \hspace{0.2cm}
$^2$ SDS, CUHKSZ \hspace{0.2cm}  
$^3$ Shenzhen Campus of SYSU \hspace{0.2cm} 
$^4$ SSE, CUHKSZ
}
\vspace{-3mm}
\maketitle

\blfootnote{Project page: \url{https://jzbao03.site/projects/PIR/}}

\maketitle 

\input{sec/1_abstract}

\input{sec/2_intro}
\input{sec/3_relatedwork}

\input{sec/4_method}

\input{sec/5_experiments}
\input{sec/6_conclusions}

{
    \small
    \bibliographystyle{ieeenat_fullname}
    \bibliography{main}
}
\input{sec/7_supp}

\end{document}

%% file: preamble.tex
\usepackage{textcomp}
\usepackage[dvipsnames]{xcolor}


%% file: sec/0_macros.tex
\usepackage{xcolor}
\usepackage{enumitem}
\usepackage{xspace}
\usepackage{booktabs}
\usepackage{colortbl}
\usepackage{multirow}
\usepackage{makecell}
\usepackage{lipsum} 
\usepackage{gensymb} 

\newcommand{\Tref}[1]{Table~\ref{#1}}
\newcommand{\eref}[1]{Eq.~\eqref{#1}}

\newcommand{\fref}[1]{Fig.~\ref{#1}}
\newcommand{\Fref}[1]{Figure~\ref{#1}}

\usepackage[labelsep=period]{caption}
\renewcommand{\paragraph}[1]{\vspace{0.2em}\noindent \textbf{#1 \hspace{0.2em}}}

\definecolor{MyDarkRed}{rgb}{0.66, 0.16, 0.16}
\definecolor{MyDarkBlue}{rgb}{0.16, 0.16, 0.66}



%% file: sec/1_abstract.tex
\begin{abstract}
This paper addresses the problem of inverse rendering from photometric images. Existing approaches for this problem suffer from the effects of self-shadows, inter-reflections, and lack of constraints on the surface reflectance, leading to inaccurate decomposition of reflectance and illumination due to the ill-posed nature of inverse rendering. In this work, we propose a new method for neural inverse rendering. Our method jointly optimizes the light source position to account for the self-shadows in images, and computes indirect illumination using a differentiable rendering layer and an importance sampling strategy. To enhance surface reflectance decomposition, we introduce a new regularization by distilling DINO features to foster accurate and consistent material decomposition. 
Extensive experiments on synthetic and real datasets demonstrate that our method outperforms the state-of-the-art methods in reflectance decomposition.
\end{abstract}

%% file: sec/2_intro.tex
\section{Introduction}
\label{sec:intro}

Inverse rendering aims to estimate the shape, materials, and lighting of a scene from 2D images. It finds applications in 3D object digitization, object manipulation, and relighting.

Recently, neural representations have achieved significant success in novel-view synthesis and 3D modeling \cite{mildenhall2020_nerf_eccv20,sitzmann2019scene,yariv2020multiview,niemeyer2020differentiable}. Neural radiance fields (NeRF), in particular, model a scene with a Multi-Layer Perceptron (MLP) that maps 3D coordinates and view directions to color and density, resulting in photorealistic rendering \cite{mildenhall2020_nerf_eccv20}. However, NeRF lacks explicit modeling of surface reflectance and lighting, making it unsuitable for relighting tasks. Several methods have been proposed to incorporate physics-based image formation models, enabling the explicit decomposition of reflectance and lighting \cite{nerv2021,physg2021}.

A branch of methods focuses on inverse rendering using images captured under environmental illumination \cite{zhang2021nerfactor,physg2021,boss2021neuralpil}. However, this presents a highly ill-posed problem due to the complex interactions between shape, materials, and lighting. Despite promising results, these methods often suffer from challenges such as the mingling of estimated surface reflectance with illumination effects, especially for real-world objects.
In a different approach, IRON \cite{iron-2022} has proposed an approach for inverse rendering from photometric images (\ie, multi-view images captured by co-locating a flashlight with a moving camera), yielding impressive results. 
Compared to environmental illumination, the flashlight (resembling a point light model) simplifies the image formation process, and the captured images contain high-frequency details (\eg, specular highlights), which are beneficial for reflectance estimation \cite{cheng2021multi, wang2020non}.

\begin{figure}[t] \centering
    \includegraphics[width=\linewidth]{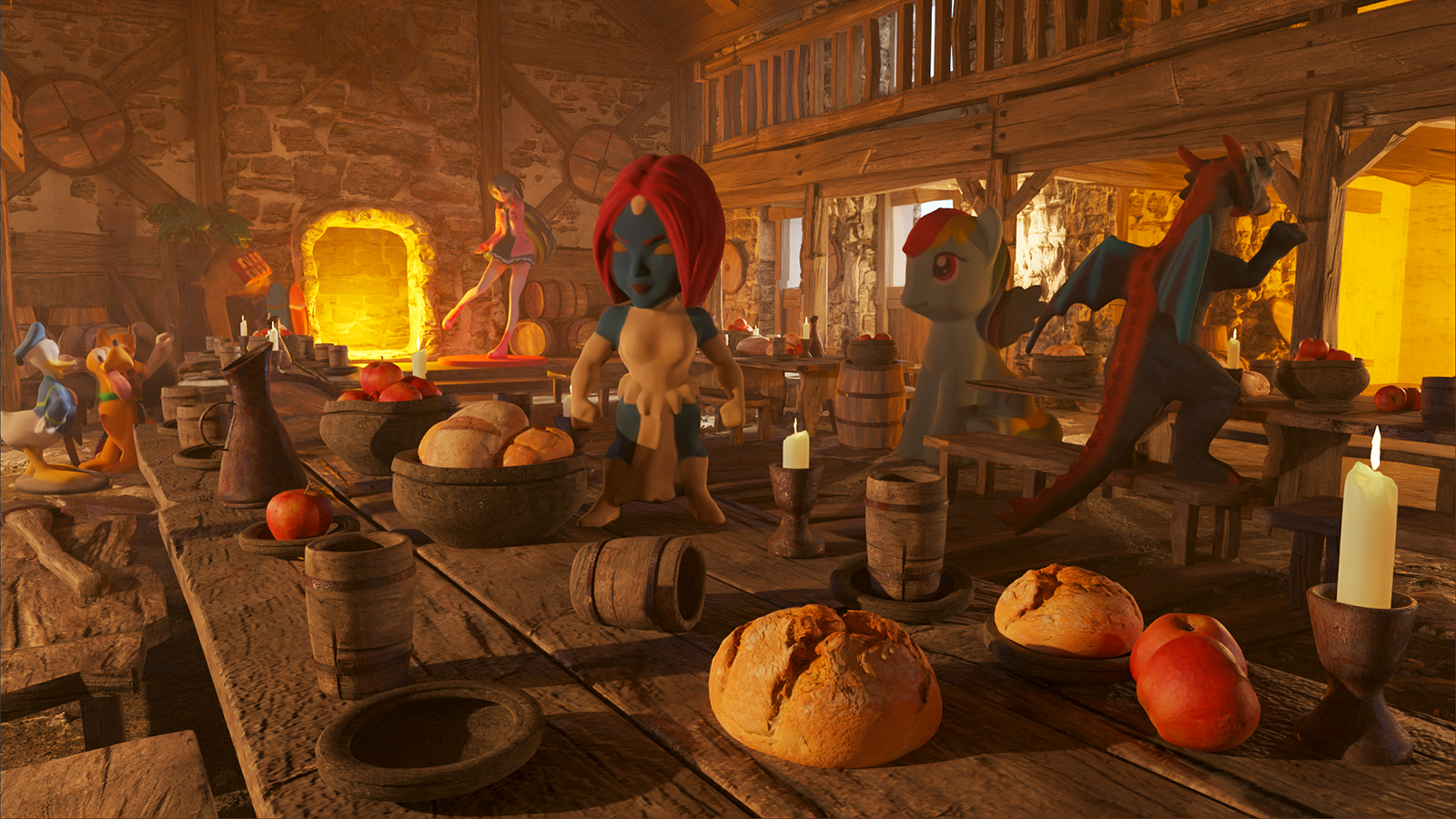}
    \caption{Reconstructed 3D assets inserted in a real game scene.} \label{fig:teaser}
    \vspace{-2em}
\end{figure}

However, IRON \cite{iron-2022} presents several shortcomings. Firstly, it assumes an ideal collocated camera-lighting arrangement, neglecting the complications posed by self-shadows that are frequently unfeasible in casual capture contexts, like smartphone use.
Secondly, it fails to consider the diverse high-frequency inter-reflections characteristic of multi-view images captured with flashlight illumination. Such oversights can lead to inaccuracies in the estimation of diffuse albedo, as it allows self-shadows to distort the outcomes or unintentionally incorporate specular highlights, particularly within concave regions. 
Additionally, the absence of effective reflectance regularization in IRON undermines the precision of material decomposition.


In this work, we introduce a novel method that leverages rich shading information available in photometric images to achieve robust inverse rendering. Notably, our approach removes the necessity of co-locating the point light source with the camera, instead opting for a joint optimization of the light source's position.
This optimization accounts for the intricate interplay between the object's geometry and the light source's position, enabling our algorithm to effectively deduce the presence of self-shadows and significantly diminish their distorting impact on the resultant images.
To accurately simulate the effect of inter-reflections, our method integrates an effective importance sampling strategy alongside a differentiable rendering layer. These techniques
effectively reducing the unwanted blending of inter-reflections on material properties.
To alleviate the ambiguity inherent in reflectance estimation,
we incorporate a DINO \cite{caron2021emerging} feature regularization into our inverse rendering framework. 
The self-supervised DINO method, by learning from extensive unlabeled datasets, captures image features encoding view-consistent contextual information across the scene, providing valuable information to understand the reflectance properties of different image regions and advancing the accuracy of the 
decomposition process.
Our key contributions are as follows:
\begin{itemize}
    \item  We propose a novel neural inverse rendering framework tailored for photometric images that jointly optimizes object shapes, materials, and lighting, achieving accurate reflectance decomposition.
    \item  We harness the shading cues present in photometric images to achieve robust inverse rendering. Our method effectively models self-shadows and employs networks alongside an importance sampling strategy for accurate inference of high-frequency inter-reflections. This approach ensures a detailed and precise rendering by capturing the subtle lighting interactions within the scene.
    \item  We introduce the DINO feature regularization for surface reflectance to group similar materials. Extensive experiments show that our method outperforms existing methods in novel view synthesis and material decomposition.
    \item  We present a new dataset containing 5 scenes captured by a mobile phone in a darkroom. The number of images per scene ranges from 120 to 400.
\end{itemize}

%% file: sec/3_relatedwork.tex
\section{Related Work}
\label{sec:related_works}

\paragraph{Neural Scene Representation}%
Neural scene representations have brought significant advancements to the fields of novel-view synthesis \cite{sitzmann2019scene,yariv2020multiview,niemeyer2020differentiable}.
The neural radiance field (NeRF) \cite{mildenhall2020_nerf_eccv20} adopts a Multi-Layer Perceptron (MLP) to represent a scene by mapping a 3D coordinate and a view direction to color and density, followed by volume rendering for pixel color computation.
To address the inherent noise in the surface derived from the density field, various efforts have leveraged the strengths of both volume rendering and surface rendering to enhance surface geometry \cite{oechsle2021unisurf,wang2021neus,yariv2021volume}.

Many follow-up methods aim to improve the performance of NeRF on different surface types. Ref-NeRF \cite{verbin2022ref} re-parameterizes NeRF’s outgoing radiance based on the reflection of the viewing vector with respect to the local normal vector, leading to improved rendering for specular surfaces \cite{ge2023ref,zhong2023color}.
Some methods extend NeRF to handle more complex scenes containing mirror surfaces \cite{tiwary2023orca,kopanas2022neural,yin2023multi,zeng2023mirror,guo2022nerfren} and transparent objects \cite{bemana2022eikonal,qiu2023looking,wang2023nemto}.

However, NeRF \cite{mildenhall2020_nerf_eccv20} lacks explicit modeling of surface reflectance and lighting, making it unsuitable for relighting tasks. In this work, we focus on performing inverse rendering from photometric images.

\paragraph{Inverse Rendering with Environment Illumination}%
A subset of methods has emerged to jointly recover the shape, materials, and lighting of objects \cite{yoshiyama2023ndjir,liang2023envidr,fan2023factored,kuang2022neroic,liang2022spidr,jin2023tensoir,sun2023joint,zhu2023efficient,mao2023neus,liu2023nero,sun2023neuralpbir} or entire scene 
\cite{wang2023neural,rudnev2022nerfosr,li2023multi,choi2022ibl,zhu2023i2,lin2024iris,yang2023complementary} using neural scene representation from multi-view images.
These methods consider an unknown distant environment illumination, and adopt diverse representations for shapes (\eg, density \cite{zhang2021nerfactor}, SDF \cite{physg2021}, and mesh \cite{Munkberg_2022_CVPR}), illuminations (\eg, spherical Gaussian \cite{boss2021nerd} and pre-integrated lighting \cite{boss2021neuralpil}), and materials \cite{mai2023neural,zhang2023nemf,baatz2022nerf,huang2023nerf,xiang2021neutex,zeltner2023real}. Recently, several studies have emerged focusing on inverse rendering through the application of 3D Gaussian splatting \cite{jiang2023gaussianshader, gao2023relightable, liang2023gs, shi2023gir}. The primary contributions of these works revolve around acceleration, which is orthogonal to our approach.

Efficiently computing inter-reflections, which involve tracing multiple bounces of rays, is a challenging problem in inverse rendering. Existing methods handle inter-reflections by assuming fixed illumination among multi-view images \cite{nerv2021,zhang2022invrender,deng2022dip,wu2023nefii,yao2022neilf,zhang2023neilf++,yang2023sireir}. For example, InvRender \cite{zhang2022invrender} introduces a MLP to map a 3D point to its indirect incoming illumination, directly derived from the outgoing radiance field. 
However, our setup involves each image being illuminated under a different point light, making the existing approach unsuitable for our scenario.

To regularize the decomposition of reflectance, NeRFactor~\cite{zhang2021nerfactor} learns the data prior for BRDFs by training an auto-encoder on the \emph{MERL dataset}~\cite{matusik2003merl}. Some methods apply low-rank or vector-quantization regularization on the reflectance \cite{zheng2023neuface,zhang2022invrender,zhong2023vq}. 
In comparison, we introduce a novel regularization without the need for additional training data by distilling the DINO \cite{caron2021emerging} feature into the object's surface. In the context of neural representation, DINO has been adopted in NeRF to scene editing ~\cite{kobayashi2022distilledfeaturefields} and grouping semantic feature~\cite{lerf2023}.

\begin{figure*}[t] \centering
    \includegraphics[width=\linewidth]{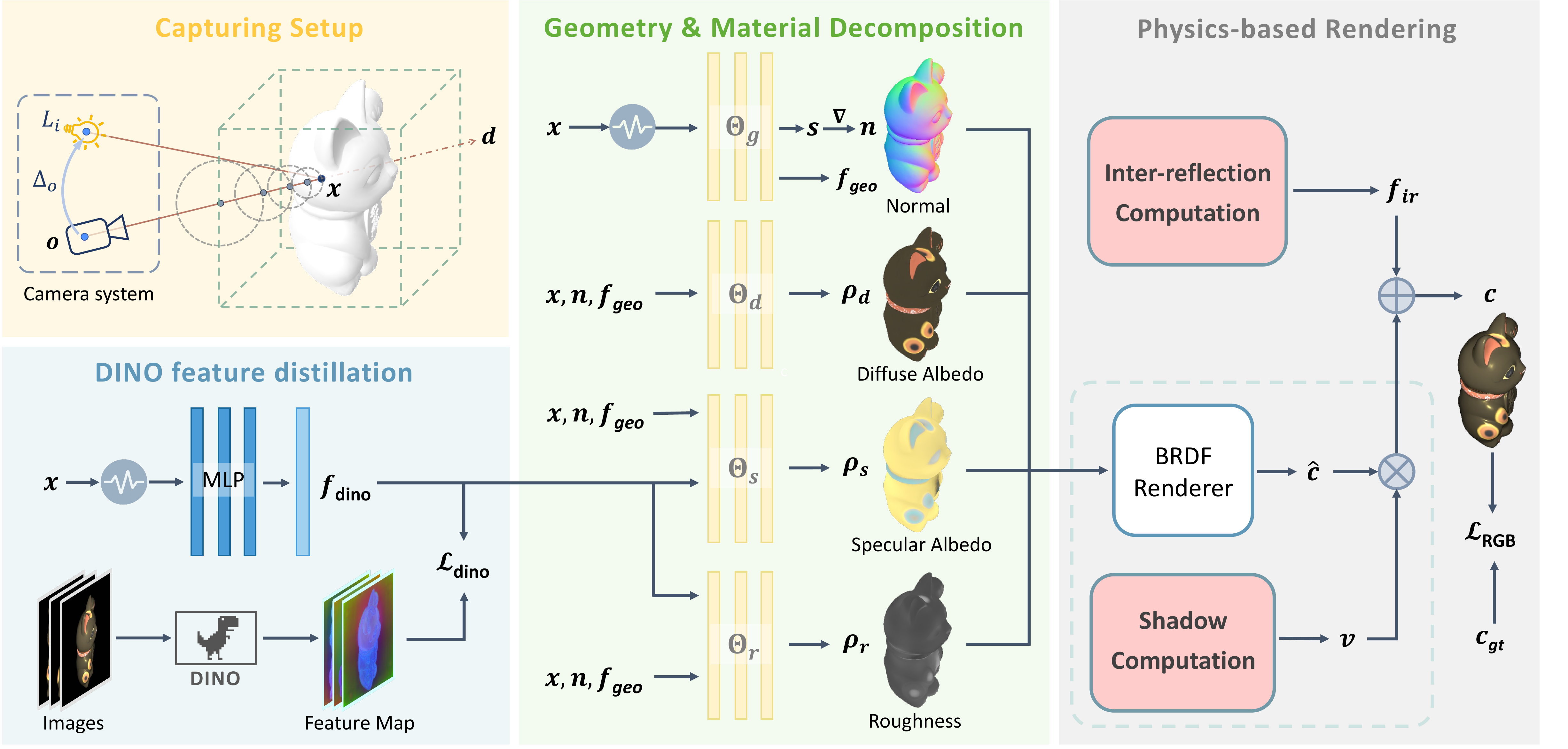}
    \caption{
    \textbf{Method overview.} Our method optimizes the light source position to account for self-shadows and model inter-reflection. 
    The DINO features are injected into the networks of specular albedo and roughness to regularize the material decomposition.} \label{fig:method}
    \vspace{-1em}
\end{figure*}

\paragraph{Inverse Rendering with Point Lights}
Different from the environment illumination, a point light model simplifies the image formation model and results in images with more high-frequency details, such as specular highlights, which significantly reduce ambiguity in inverse rendering \cite{zeng2023nrhints,xu2023renerf,zhu2023neural,toschi2023relight,liu2023openillumination}.
Several methods have been proposed to improve the accuracy of inverse rendering by utilizing a point light model \cite{bi2020deep,bi2020neural,yang2022psnerf,guo2022edge,brahimi2022supervol,cheng2023wildlight,li2023neural}.
IRON \cite{iron-2022} employs the NeuS method \cite{wang2021neus} to represent the surface and utilizes an edge-aware physics-based surface rendering for geometry refinement and materials estimation. However, IRON assumes an ideal collocated camera-lighting setup and overlooks self-shadows and inter-reflections. In contrast, our method addresses all these issues and explicitly regularizes surface reflectance to achieve more accurate inverse rendering.
While some methods compute shadows during optimization, they typically assume the environment illumination \cite{zhang2021nerfactor,chen2022tracing,verbin2023eclipse} or known point light positions \cite{yang2022s3nerf,ling2023shadowneus,liu2023whatfromashadow,Tiwary2022learningneuralshadows}. Differently, our method utilizes self-shadow cues to calibrate the light positions.



%% file: sec/4_method.tex
\newcommand{\viewdir}{\boldsymbol{d}}
\newcommand{\pointcolor}{\boldsymbol{c}}
\newcommand{\density}{\sigma}
\newcommand{\mat}{\rho}
\newcommand{\albedo}{d}
\newcommand{\salbedo}{s}
\newcommand{\rough}{r}
\newcommand{\envmap}{\boldsymbol{L}_{SG}}
\newcommand{\ray}{\boldsymbol{r}}
\newcommand{\pixelcolor}{\boldsymbol{C}}
\newcommand{\coarse}{{(c)}}
\newcommand{\refine}{{(r)}}
\newcommand{\loss}{\mathcal{L}}
\newcommand{\gt}{\mathrm{(gt)}}
\newcommand{\uncertain}{{(\tau)}}
\newcommand{\point}{\boldsymbol{x}}
\newcommand{\img}[2]{I_{#1}^{#2}}
\newcommand{\nview}{M}
\newcommand{\viewindx}{N_p}
\newcommand{\nlight}{L_m}
\newcommand{\imgset}{\mathcal{I}}
\newcommand{\camloc}{\boldsymbol{o}}

\newcommand*\diff{\mathop{}\!\mathrm{d}}
\newcommand*\Diff[1]{\mathop{}\!\mathrm{d^#1}}
\newcommand{\lout}{\hat{I}}
\newcommand{\lin}{L_i}
\newcommand{\lind}{L_\text{ind}}
\newcommand{\din}{\boldsymbol{w}_i}
\newcommand{\dout}{\boldsymbol{w}_o}
\newcommand{\dref}{\boldsymbol{w}_r}
\newcommand{\normal}{\boldsymbol{n}}
\newcommand{\residual}{f_{ir}}
\newcommand{\vis}{n}
\newcommand{\brdf}{f_r}
\newcommand{\nmlp}{f_n}
\newcommand{\vmlp}{f_v}
\newcommand{\surf}{\mathcal{S}}
\newcommand{\surfpts}{\mathcal{S}_{\boldsymbol{x}}}
\newcommand{\surfnorm}{\mathcal{N}_{\sigma}}
\newcommand{\surfvis}{\mathcal{V}_{\sigma}}
\newcommand{\psnormalmap}{\mathcal{N}_{m}}
\newcommand{\R}{\mathcal{R}}

\newcommand{\latent}{\boldsymbol{z}}
\newcommand{\fgeo}{\boldsymbol{f}_{\text{geo}}}
\newcommand{\fdino}{\boldsymbol{f}_{\text{dino}}}
\newcommand{\pixel}{\boldsymbol{p}}
\newcommand{\SDF}{S_{\boldsymbol{\Theta}_g}}

\section{Method}

\subsection{Overview}

\paragraph{Capturing Setting}
In this work, we focus on reconstructing object geometries, materials, and illumination conditions from multi-view images lit by a flashlight. Previous methods \cite{iron-2022, cheng2023wildlight} assume a collocated camera-light setup and optimize the scene with a simplified rendering model. Such an ideal setting is impractical to attain in our daily capture, like a mobile phone; we consider a more general fixed setup akin to a camera mounted on a mobile phone. Our method (\fref{fig:method}) tackles the complexities introduced by self-shadows, inter-reflections, and ambiguities in reflectance estimation, which are prevalent issues in current techniques.



\paragraph{Rendering Equation}
In theory, the rendering equation~\cite{kajiya1986rendering} for a surface point $\point$ can be written as
\begin{equation}
    \label{eq:render_eq_general}
    \lout (\dout; \point) = \int_\Omega \lin (\din; \point) \brdf (\dout, \din; \point) (\din \cdot \normal) \diff \din,
\end{equation}
where $\lin (\din; \point)$ denotes the incoming radiance arriving from direction $\din$, and $\brdf (\dout, \din; \point)$ encapsulates the surface's bidirectional reflectance distribution function (BRDF) at $\point$. 
This equation calculates the outgoing radiance $\lout (\dout; \point)$ of point $\point$ in the direction of $\dout$ by integrating all radiance contributions over the upper-hemisphere $\Omega$ surrounding the surface normal $\normal$.

By assuming a point light and considering light visibility and indirect lights, the rendering can be approximated as
\begin{equation}
\begin{aligned}
    \label{eq:render_eq_ours}
    \lout (\dout; \point) = \lin (\din; \point) \brdf (\dout, \din; \point) (\din \cdot \normal) \\
    \times \vmlp (\din; \point) + \residual(\dout; \point),
\end{aligned}
\end{equation}
where $\vmlp (\din; \point)$ is the visibility of light along $\din$ at $\point$ that models self-shadows in the rendered image, and $\residual$ accounts for the residual effects attributed to inter-reflections.

\paragraph{Pipeline}
Our pipeline commences with estimating the object's geometry and surface diffuse albedo using the off-the-shelf neural surface reconstruction framework, NeuS \cite{mao2023neus}. Subsequently, we utilize physics-based rendering to jointly refine the geometry and materials of the object as well as the position and intensity of the flashlight. Our approach leverages differentiable rendering techniques to accurately model self-shadows and indirect illumination in photometric images, thereby achieving robust material decomposition. Additionally, we reduce the ambiguities of surface reflectance decomposition by integrating self-supervised DINO \cite{caron2021emerging} features from multi-view images. Our method can export the mesh and texture maps of the optimized 3D models, which can be seamlessly integrated into conventional rendering pipelines, as shown in~\fref{fig:teaser}.


\subsection{Neural Scene Representation}


\paragraph{Geometry Representation}
In our approach to representing scene geometry, the geometry is represented by the zero level set of a SDF $\mathcal{S}=\left\{\point \in \mathbb{R}^3 \mid s(\point)=0\right\}$ in line with recent advancements in the field \cite{wang2021neus,yariv2020multiview,zhang2022invrender}. For any point $\point \in \mathbb{R}^3$, the signed distance $s$ and a learned local geometric feature descriptor of $\point$ are parameterized by a MLP, denoted as $f_{\Theta_g} = (s, \fgeo) \in \mathbb{R} \times \mathbb{R}^{256}$.

Color rendering for a pixel is achieved through the projection of a ray $\mathbf{r}(t) = \mathbf{o} + t\mathbf{d}$  from the camera's origin $\mathbf{o}$, extending in the view direction $\mathbf{d}$. In the volumetric field, the color $C$ rendered from a pixel is obtained by integrating along its ray path, with the integral approximated over $N$ discrete points as follows:
\begin{equation}
    \label{eq:radiance_field}
    C(\mathbf{o}, \mathbf{d}) =\sum_{j=1}^N T_j\left(1-\exp \left(-\sigma_j \delta_j\right)\right) \pointcolor_j,
\end{equation}
where $T_j =\exp \left(-\sum_{q=1}^{j-1} \sigma_q \delta_q\right)$ denotes the accumulated transmittance at sampled point $\mathbf{r}(t_{j})$, and $\pointcolor_j$ represents the point's color.
We incorporate the unbiased density conversion method \cite{mao2023neus}, translating SDF values into density representations for the scene's geometry.

\paragraph{Materials Representation}
To achieve physics-based rendering, our framework decomposes the scene's BRDF into diffuse and specular components, utilizing the roughplastic model for microfacet specular reflection \cite{walter2007microfacet}.
The materials at a point $\boldsymbol{x}$ include the diffuse albedo $\mat_\albedo$, specular albedo $\mat_\salbedo$, and roughness $\mat_\rough$. 
These spatially-varying BRDF parameters are encapsulated by MLPs with a positional encoding function and optimizable parameters $\Theta_{ \albedo }$, $\Theta_{ \salbedo }$, $\Theta_{ \rough }$. 

\subsection{Light Source Optimization}
\paragraph{Lighting Model} The flashlight is modeled as a white point light source, similar to the configurations in \cite{iron-2022,cheng2021multi}. We assume the photographic equipment is in a rigid capture setup, such that the relative positions of the camera and point lights are fixed across images.

Denoting the relative offset as a learnable parameter $\Delta_o$, the incident light direction $\din(\point)$ and intensity $L_i(\din; \point)$ for a surface point $\point$ are defined as
\begin{align}
    \label{eq:light_fall_off}
    \din(\point) = \frac{(\camloc + \Delta_o) - \point}{\|(\camloc + \Delta_o) - \point\|}, 
    L_i(\din; \point) = \frac{L}{\| (\camloc + \Delta_o) - \point\|^2},
\end{align}
where $\camloc$ is the camera center, and $L$ represents the learnable scalar intensity of the light.

\paragraph{Visibility Computation} By leveraging the object's geometry and the point light's position, we infer self-shadows and mitigate their effects on the captured images.
To make the process differentiable, we sample $N$ points along a direction from the point $\point$ to the light position, denoted as $\din$, and calculate the visibility of a point as \cite{zhang2021nerfactor}:
\begin{align}
    \label{eq:visibility}
    \vmlp (\din; \point) = 1 - \sum_{j=1}^{N} \alpha_{j} \prod_{k=1}^{j-1}\left(1-\alpha_k\right),
\end{align}
where $\alpha_j$ is the discrete opacity value. We compute the visibility online so that the object geometry and light position can be jointly optimized.

\subsection{Differentiable Inter-reflection Computation}




Prior methods, like InvRender~\cite{zhang2022invrender}, sample multiple rays and use an MLP to cache the indirect incoming illumination at a surface point as smooth SGs under a static illumination, 
hindering their application in scenes dynamically captured under directional lighting, such as with a flashlight.

To address the issues of missing indirect illumination details and the high computational load, we propose an online indirect illumination computation strategy based on importance sampling (see supp. for pipeline details). 

\paragraph{Importance Sampling}
Inter-reflection occurs when light reflects from one surface to another. We observe that the specular surfaces exhibit more pronounced inter-reflections, and the main source of indirect illumination for a surface point $\point$ comes from its reflective direction $\dref$, which is the reflection of the view angle around the surface normal. To efficiently compute the indirect light, we sample multiple rays near $\dref$ for indirect radiance calculation. 
Initially, we identify the secondary intersection point $\point'$ where the indirect bounce meets the surface, and then we compute the incoming radiance $\lind(\din; \point)$ as the outgoing radiance from $\point'$. Radiance rendering of the secondary intersection point only takes into account the intense lighting from the flashlight and $\point'$, excluding points occluded from the flashlight. The indirect illumination results from integrating all these incoming radiances over the upper hemisphere around $\point$ surface normal.

To mitigate the artifacts of insufficient sampling, we blend the radiance from the incoming direction $\din$ around the reflective view using a learnable scalar $\gamma$. One straightforward parameterization approach involves mapping the scalar from the point coordinates and the implicit geometric feature. However, we discovered that relying on point position and local geometry information often restricts the representation of varying inter-reflections, particularly in concave areas. 
A more effective strategy involves utilizing common physical properties (distance, view direction, and roughness) to deduce dynamic indirect illumination. Consequently, we express the indirect illumination component $\residual(\point)$ as:
\begin{align}
    \residual(\point) = \gamma \cdot \sum_{\din} \lind(\din; \point) \brdf(\din, \dout; \point) (\din \cdot \normal).
\end{align}

\subsection{DINO Regularization}
To reduce the inherent ambiguity of reflectance estimation, 
we introduce a novel reflectance regularization based on the distilled DINO feature field. 
DINO \cite{caron2021emerging} displays inherent capabilities for object decomposition by training on diverse unlabeled data and has been successfully distilled into 3D fields for radiance editing \cite{kobayashi2022distilledfeaturefields} and open-vocabulary object grouping \cite{lerf2023}. Inspired by these methods, we propose to distill the DINO feature from 2D images to 3D surfaces (object geometry) to learn a fine composition and contextual information of the object, resulting in a more consistent decomposition of surface reflectance and materials. In our implementation, we distill the DINO feature to the initial geometry field by minimizing the loss function:
\begin{equation}
    \label{eq:dino}
     \loss_{\text{dino}} = \sum_{\pixel} (\fdino(\point(\pixel)) - \text{DINO}(\pixel)) ^ 2,
\end{equation}
where $\pixel$ denotes the pixel in the 2D images, $\point(\pixel)$ indicates the corresponding 3D surface point derived by ray tracing. The distillation process minimizes the square distance between the learnable DINO feature $\fdino$ on the surface and the ViTs pre-trained on 2D images. The distilled DINO features are incorporated into the networks of specular albedo and roughness, providing regularization to enhance the accuracy of material decomposition.

In our experiment, we found that the resolution of the DINO feature also influences the ability to distinguish objects' composition. Higher resolution DINO features can assist in achieving finer material decoupling. 
Empirically, we upsample the image by a scaling factor of two to extract DINO features with a higher resolution.

%

\subsection{Optimization}

\paragraph{Differentiable surface point}
To make the surface point differentiable, we reparameterize the surface intersection equation as previous works \cite{yariv2020multiview, iron-2022}:
\begin{align}
    \boldsymbol{x}_{\boldsymbol{\Theta}_g}=\boldsymbol{x}-\frac{\boldsymbol{n}}{\boldsymbol{n}^T \boldsymbol{n}} S_{\boldsymbol{\Theta}_g}(\boldsymbol{x})=\boldsymbol{x}-\boldsymbol{n} S_{\boldsymbol{\Theta}_g}(\boldsymbol{x}),
\end{align}
where $\SDF(\point)$ denotes the SDF value of point $\point$, $\normal$ is the normal vector at $\point$ calculated by $\normal = \nabla \SDF(\point)$.

\paragraph{Training Loss}
The optimization process is formulated as a minimization problem where the total loss $\loss$ is a combination of several components, each targeting a specific aspect of the reconstruction:
\begin{equation}
    \label{eq:optim}
    \loss = \loss_\text{rgb} + \loss_\text{ssim} + \lambda_1 \loss_\text{eik} + \lambda_2\loss_{\alpha} + \lambda_3 \loss_\text{smooth} + \lambda_4 \loss_\text{dino}.
\end{equation}
$\loss_\text{rgb}$ is the $L_2$ loss computed on the Gaussian pyramids of the predicted image $\hat{I}$ and the reference image $I$. 
$\loss_\text{SSIM}$ is the SSIM loss \cite{wang2004image}. 
$\loss_\text{eik}$ is the Eikonal loss \cite{gropp2020implicit} to regularize the MLP for a valid SDF. 
$\loss_{\alpha}$ is the roughness range loss, set at 0.5. 
The first four loss terms are the same as IRON \cite{iron-2022}. 
$\loss_\text{smooth}$ is the smoothness loss on the specular albedo and roughness as used by \cite{yao2022neilf}. 
$\loss_\text{dino}$ denotes the DINO feature alignment loss described in \eref{eq:dino}.
During the inverse rendering stage, the edge-aware surface rendering proposed by IRON is adopted to refine the geometry \cite{iron-2022}.

%% file: sec/5_experiments.tex
\section{Experiments}

\begin{figure}[t] \begin{center}
    \includegraphics[width=0.9\linewidth]{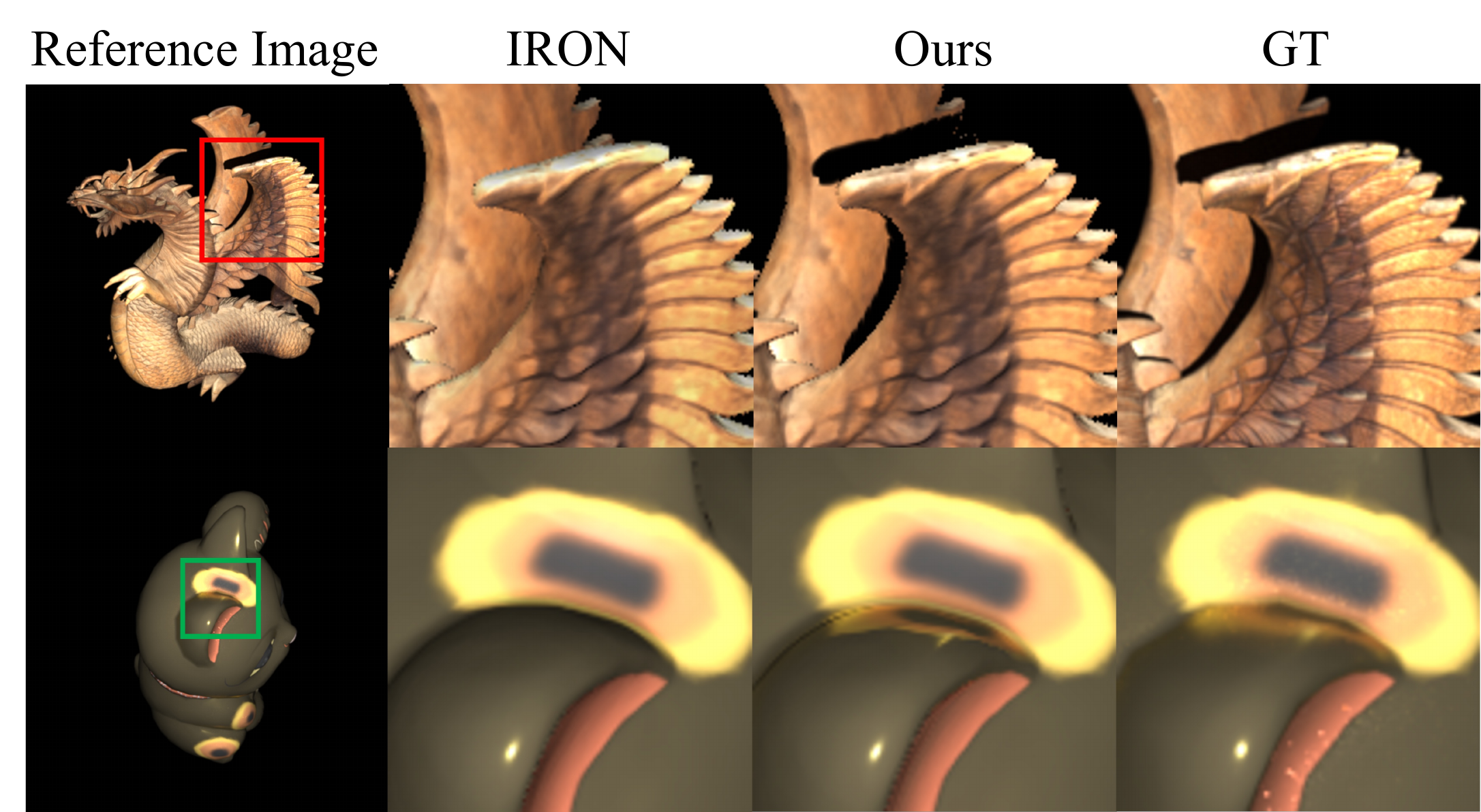}
    \\
    \caption{Visual results of self-shadows and inter-reflections.}
    \label{fig:res-shadow-inter}
    \vspace{-2.5em}
\end{center} \end{figure}

\subsection{Datasets} 


\paragraph{Synthetic Data}
The synthetic dataset comprises six objects. 
Four objects with a variety of shapes and materials, namely \emph{duck}, \emph{maneki}, \emph{horse}, and \emph{dragon}, are used in IRON \cite{iron-2022}. 
To conduct a thorough analysis,
we adopt another two objects: \emph{marble bowl}, a concave bowl with complex light effects, and \emph{armchair} with self-shadows in multiple views. 

We consider a practical non-collocated flashlight and camera setting, similar to a mobile phone setup, with the angle between the camera and flashlight to the object center set to about 3 degrees.
We render 200 images from random views under a non-collocated flashlight via Mitsuba \cite{jakob2010mitsuba} for training. We also render 100 images together with their diffuse albedo maps, specular albedo maps, and roughness maps for test images to evaluate the quality of novel view synthesis and material decomposition. 

\paragraph{Real Data} We tested our method on the DRV dataset \cite{bi2020deep} captured by a nearly collocated camera-flashlight setup, and the Luan dataset \cite{luan2021unified} captured using a smartphone. 
We also captured a dataset by an iPhone in a darkroom.
Camera poses for real images were derived using COLMAP \cite{schonberger2016_sfm_cvpr16}.


\begin{figure*}[t] \begin{center}
    \captionof{table}{Quantitative comparison of novel view rendering results with other inverse rendering methods on the synthetic dataset.} 
    \label{tab:res_offset_synthetic}
    \input{tables/res_offset_synthetic}

    \includegraphics[width=\linewidth]{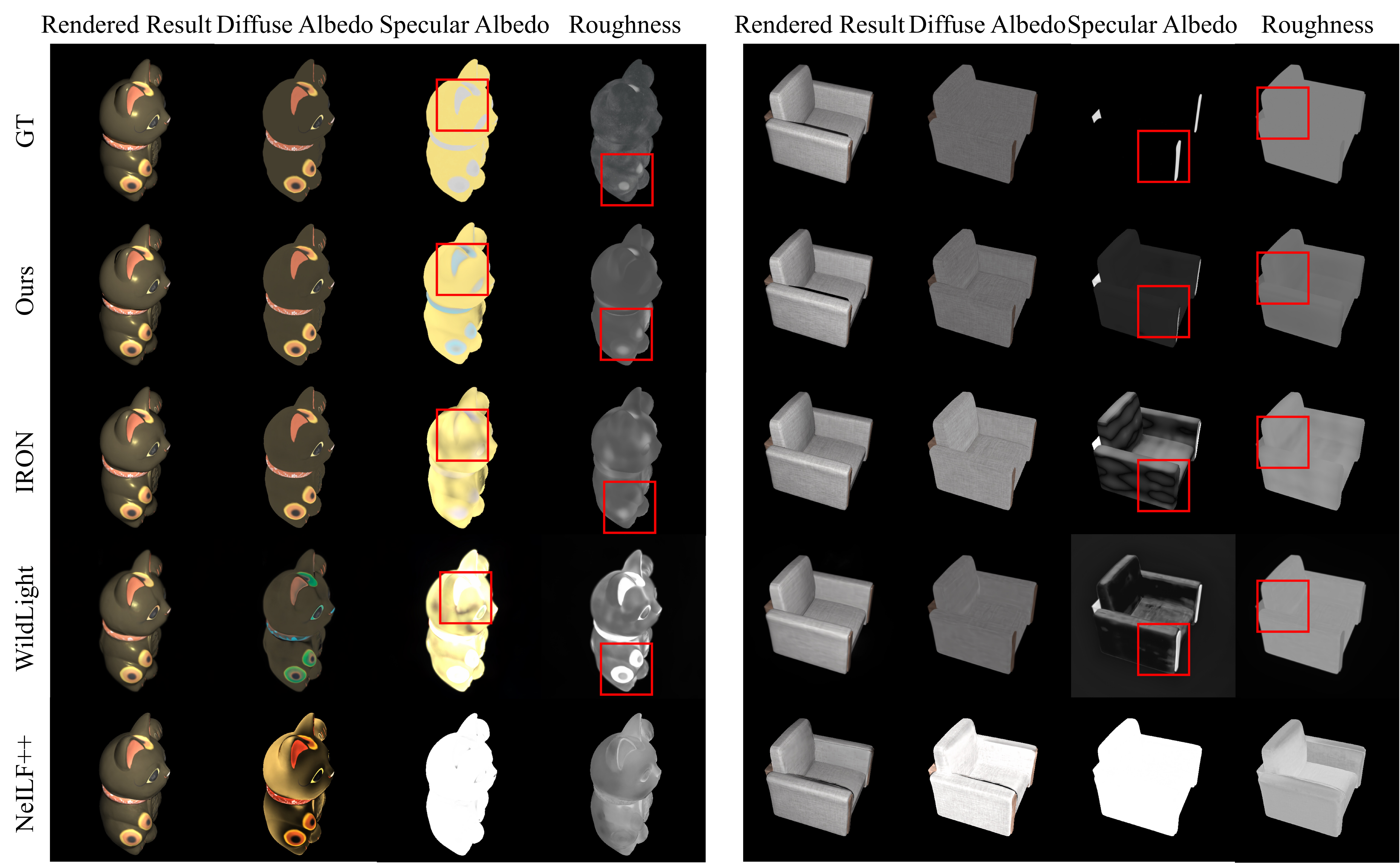}
    \captionof{figure}{\textbf{Qualitative comparison of state-of-the-art methods and our method on the synthetic dataset.} The materials of NeILF++\cite{zhang2023neilf++} are \emph{Base Color}, \emph{Metallic}, \emph{Roughness} defined by simplified Disney principled BRDF and others are using Mitsuba roughplastic model.}
    \label{fig:res_synthetic}
    \vspace{-2em}
\end{center} \end{figure*}

\begin{table*}[t] \begin{center}
    \captionof{table}{\textbf{Quantitative comparison on synthetic data.} The predicted albedos are scaled to match the GT light intensity during evaluation. 
    }
    \label{tab:res_material_estimation}
    \input{tables/res_material_estimation}

    \vspace{-0.1em}
    \includegraphics[width=\linewidth]{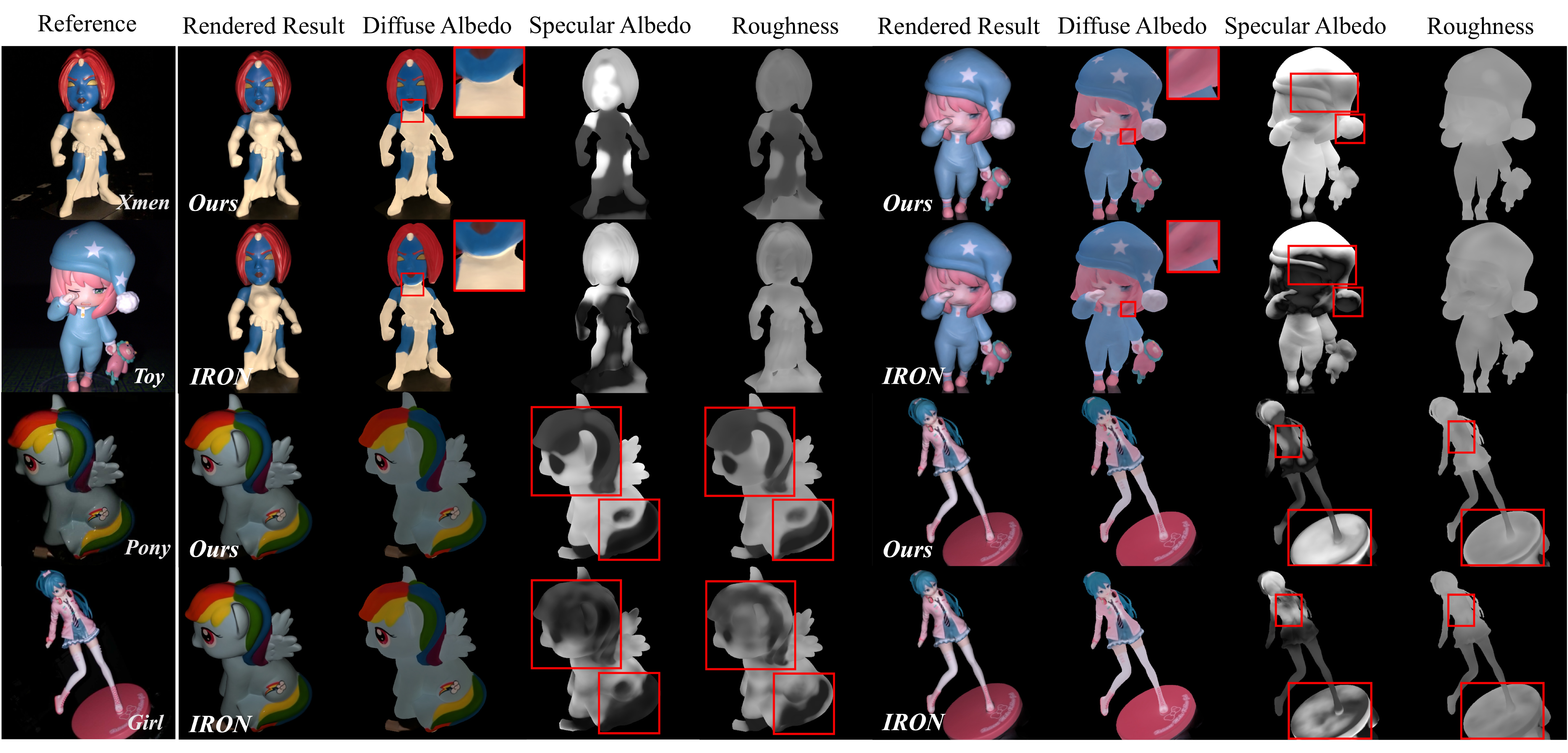}
    \vspace{-2em}
    \captionof{figure}{
    \textbf{Visual results of material decomposition on real data.}
    For each object, we compare the results of our method and IRON. \emph{Xmen} is from Luan dataset \cite{luan2021unified}, \emph{Toy} is a self-captured dataset, \emph{Pony} and \emph{Girl} are from the DRV dataset \cite{bi2020deep}. }
    \label{fig:res_real}
    \vspace{-2em}
\end{center} \end{table*}

\subsection{Comparison with Existing Methods}


For a fair comparison of material decomposition, we adapt the physical shader in WildLight to roughplastic model \cite{walter2007microfacet}, as utilized by Mitsuba. We then integrate the decomposed material components along the rays to generate the material maps of the view.
Comparison with the methods on volume rendering method DRV \cite{bi2020deep} and the mesh-based approach PSDR \cite{luan2021unified} were not conducted in our study, as their codes are not available.
We also compared with other implicit methods presented by \cite{zhang2023neilf++}, which recover neural fields while considering inter-reflections.

Our method can recover sharp inter-reflection details and complex self-shadow caused by non-collocated camera and flashlight (see \fref{fig:res-shadow-inter}),
resulting in more accurate specular albedo and roughness (see \fref{fig:res_synthetic}). 
Existing methods for the similar inverse rendering settings (\ie, IRON~\cite{iron-2022} and WildLight~\cite{cheng2023wildlight}) overlook inter-reflections and self-shadows, leading to inaccuracy in material recovering. Specifically, the diffuse albedo often blends indirect illumination, particularly in concave areas. Self-shadows distort surface reflectance, leading to incorrect specular albedo brightness and noisy roughness. 
The state-of-the-art implicit method NeILF++ \cite{zhang2023neilf++} tends to erroneously blend the intensity of moving light sources into material properties.

\Tref{tab:res_offset_synthetic} and \Tref{tab:res_material_estimation} show the quantitative comparison of rendering and material decomposition, respectively. We can see that our method achieves more accurate results, especially in the estimation of specular albedo and roughness.

\subsection{Results on Real Data}

We compare our method with IRON on the challenging real dataset. 
\Fref{fig:res_real} showcases the rendered images and material decomposition results. Compared with IRON, our method exhibits fewer shadows and indirect illumination effects baked into the diffuse albedo, giving credit to our modeling of inter-reflection and lighting optimization. 

\begin{figure}[t]
\vspace{-.5em}
\begin{center}
    \includegraphics[width=.96\linewidth]{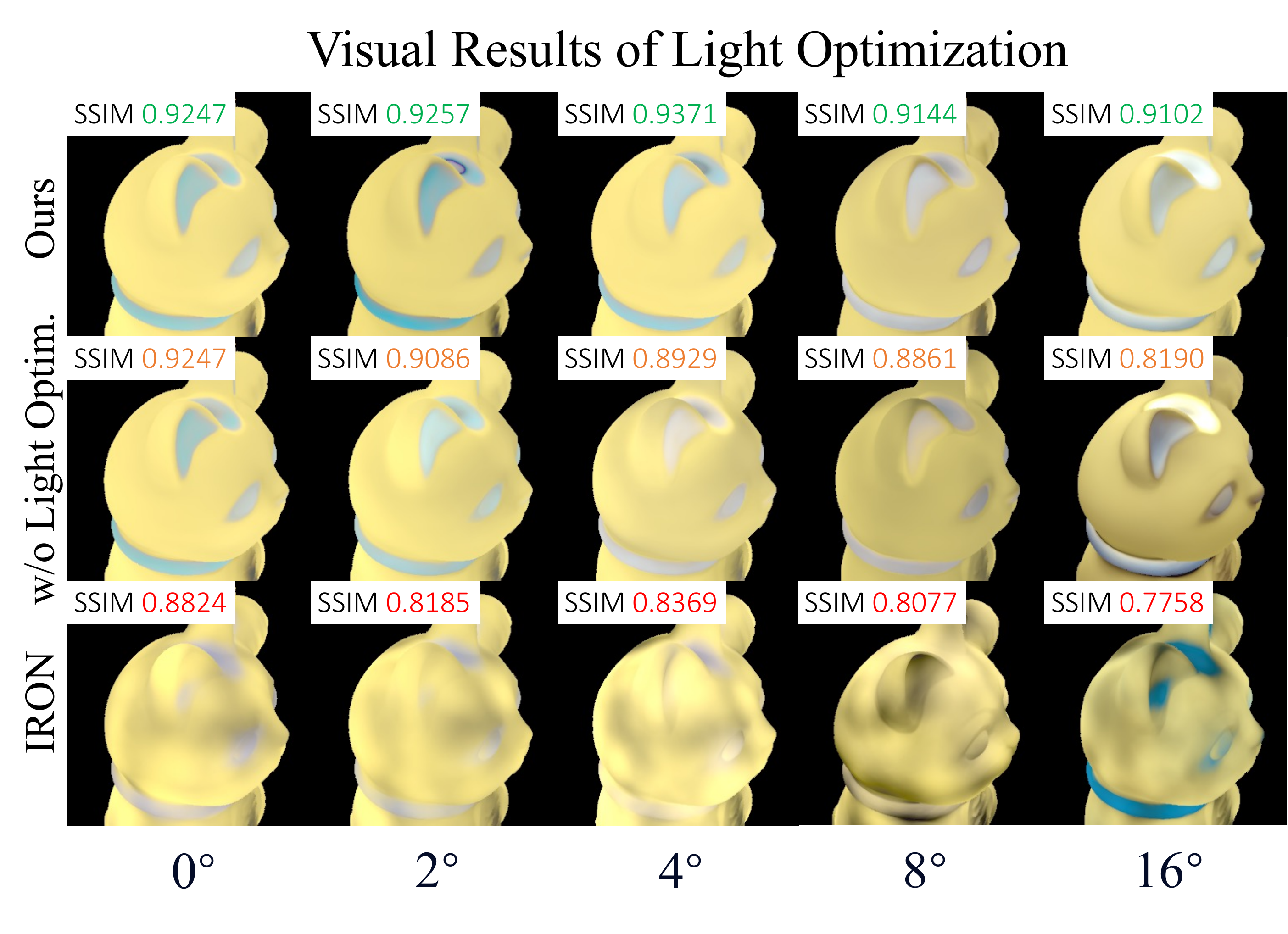}
    \vspace{-0.5em}
    \caption{Quantitative and visual results of light optimization.}
    \label{fig:fig-ablation-light}
    \vspace{-2em}
\end{center}
\end{figure}


\begin{figure}[t] \begin{center}
    \vspace{-1em}
    \includegraphics[width=\linewidth]{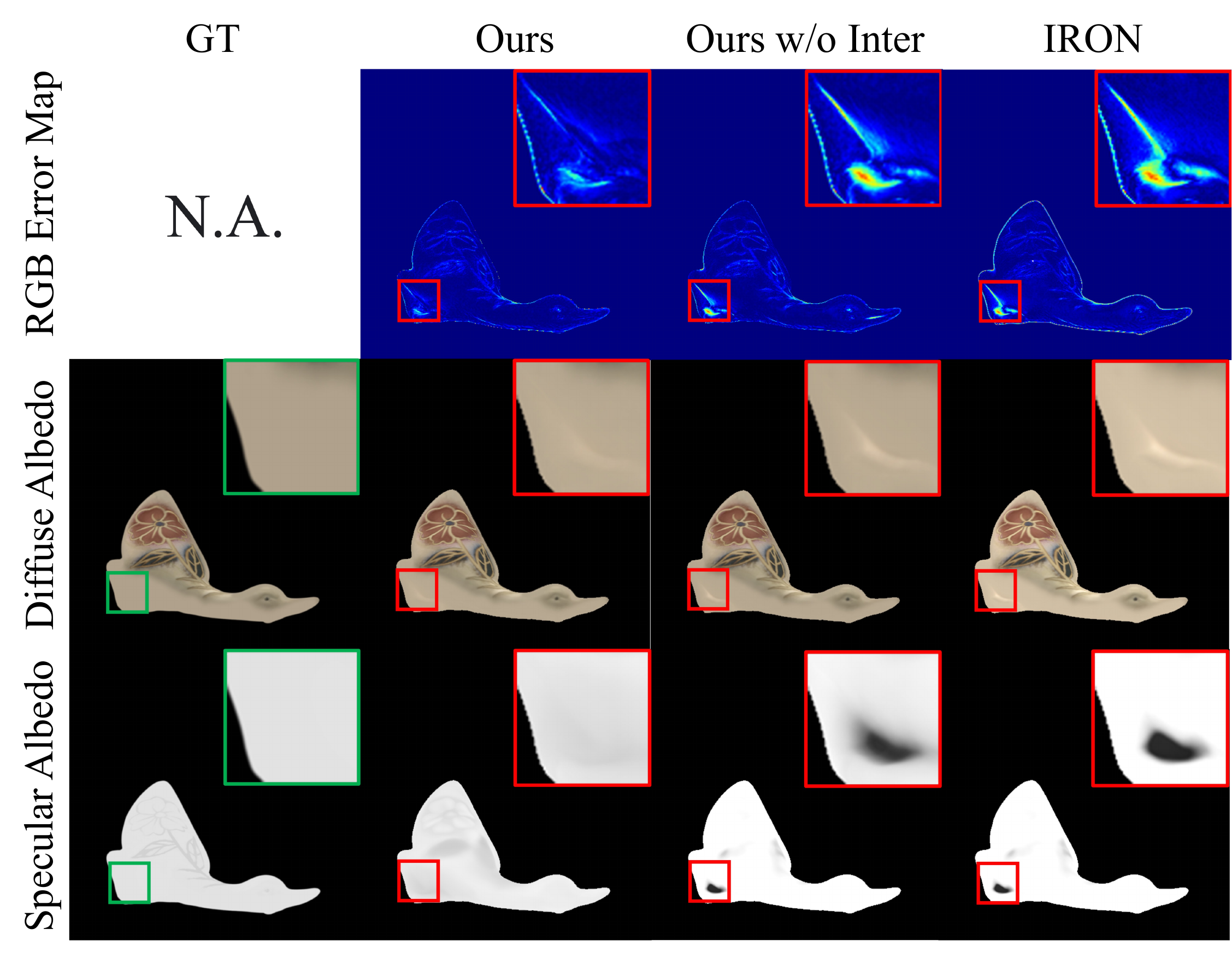}
    \caption{\textbf{Ablation study on inter-reflection.} We render scenes in simplified \emph{collocated setting} without self-shadows.}
    \label{fig:ablation-inter}
    \vspace{-3em}
\end{center} \end{figure}

Specifically, in the diffuse albedo of \emph{Xmen} estimated by IRON, the neck region bakes indirect illumination and appears brighter. 
In IRON's result on \emph{Toy}, self-shadows distort the diffuse albedo, embedding shadows within the material. With the DINO regularization, our method produces more consistent reflectance decomposition (see \emph{Pony} and \emph{Girl}).

\subsection{Ablation Studies}
To gain a deeper insight into the efficacy of our approach, we have conducted a thorough analysis of our method. We evaluate the inter-reflection modeling, lighting optimization, and DINO regularization to validate our method.

\paragraph{Evaluation on Lighting Optimization}
We show the strength of our method in calibrating the camera-lighting offset even in the extreme case. 
The quantitative and qualitative comparison in \fref{fig:fig-ablation-light} shows that, without the light position optimization, the method fails to accurately estimate materials with large light source deviation.



\begin{table*}[t] \begin{center}
    \captionof{table}{Quantitative comparison of material estimation on synthetic dataset under collocated camera-lighting (collocated setup).} 
    \label{tab:res_collocated_synthetic}
    \input{tables/res_collocated_synthetic_material}

    \vspace{0.5em}
    \includegraphics[width=\textwidth]{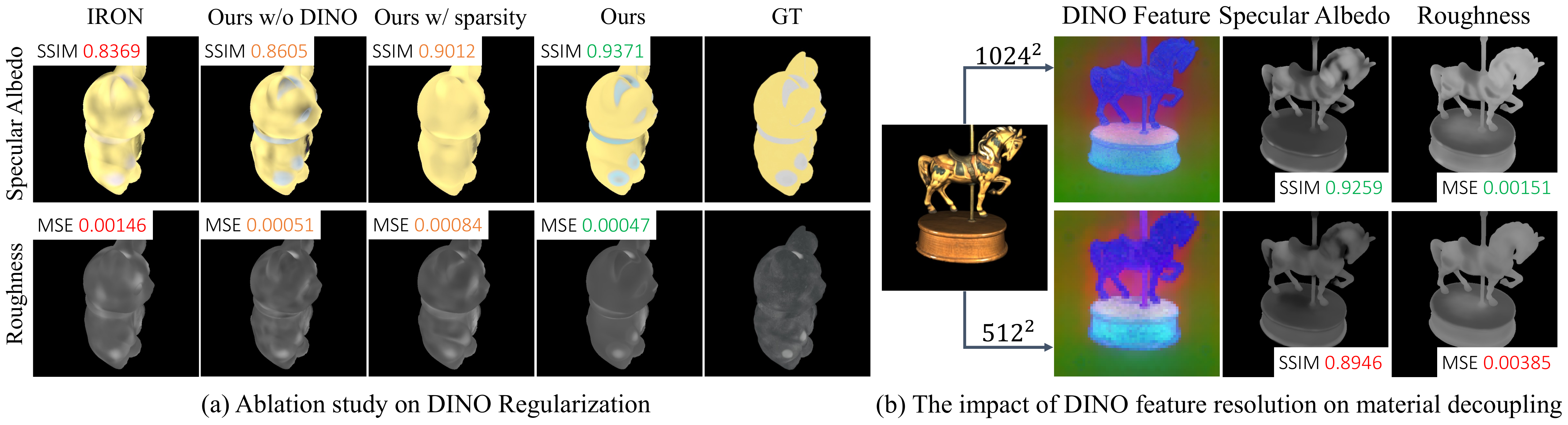}
    \\
    \vspace{-0.5em}
    \captionof{figure}{Ablation study on DINO feature regularization (a) and the impact of DINO resolutions on material decoupling (b).}
    \label{fig:res_dino}
    \vspace{-1.5em}
\end{center} \end{table*}

\paragraph{Evaluation on Inter-reflection Modeling} 
The synthetic data rendered for the ablation study on inter-reflection is in a collocated camera-lighting setting to avoid the influence of self-shadows and different physical shader settings used in different methods.
We ablate the inter-reflection calculation and compare the results in \Tref{tab:res_collocated_synthetic} and \fref{fig:ablation-inter}. 
Our method accurately estimates the materials, and without the inter-reflection modeling, the predicted specular albedo has artifacts, and diffuse albedo baked the indirect illumination.

\begin{figure}[t] \begin{center}
    \vspace{-1.4em}
    \includegraphics[width=\linewidth]{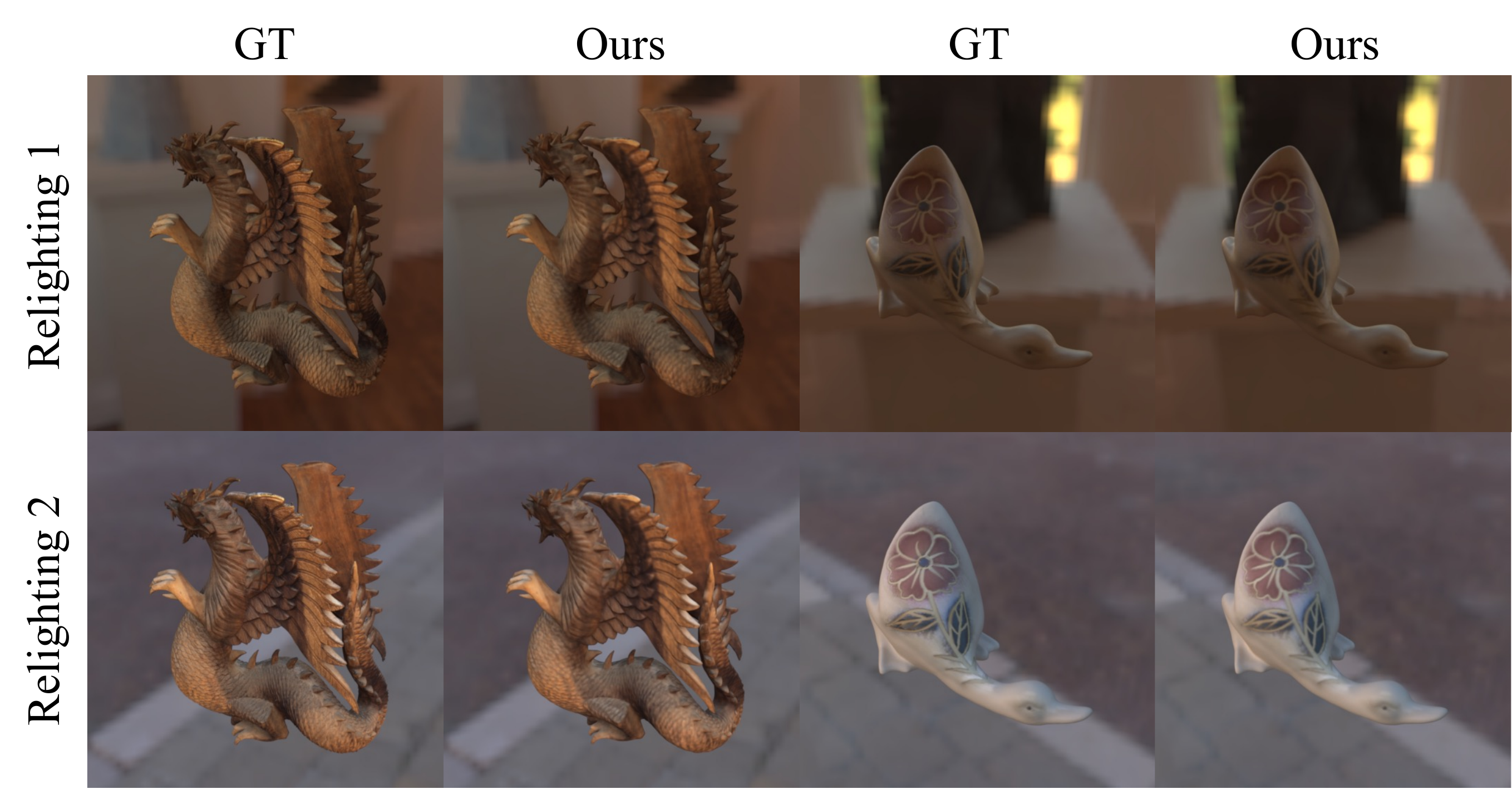}
    \caption{Relighting results with the estimated materials.}
    \label{fig:res_relighting}
    \vspace{-2.6em}
\end{center} \end{figure}

\paragraph{Evaluation on DINO regularization}
Similarly, we evaluate the DINO regularization under the collocated camera-lighting setting for the ablation study. \Tref{tab:res_collocated_synthetic} and \Fref{fig:res_dino}~(a) shows the quantitative and visual results respectively.
Our decomposition results surpass other methods and we also show the DINO regularization is much better than some empirical regularization used in \cite{zhang2022invrender, deng2022dip}. 
The DINO feature regularization can ease the inherent difficulty of reflectance decomposition by grouping consistent, contextual information across the scene. 


In addition, \fref{fig:res_dino}~(b) shows the influence of DINO feature resolution in reflectance decomposition, validating that higher resolution of DINO features can assist in achieving finer material decoupling.





\subsection{Relighting Results}

We relight the objects with estimated material properties under two environments and show results in \fref{fig:res_relighting}. This highlights our method's ability to precisely recover material properties, thereby enabling further relighting applications.



%% file: tables/res_offset_synthetic.tex
\resizebox{\textwidth}{!}{
\begin{tabular}{l||*{3}{c}|*{3}{c}|*{3}{c}|*{3}{c}|*{3}{c}}
    \toprule
    & \multicolumn{3}{c|}{\emph{Duck}} 
    & \multicolumn{3}{c|}{\emph{Maneki}}
    & \multicolumn{3}{c|}{\emph{Horse}}
    & \multicolumn{3}{c|}{\emph{Marble Bowl}}
    & \multicolumn{3}{c}{\emph{Dragon}} \\
    Method & PSNR & SSIM & LPIPS & PSNR & SSIM & LPIPS & PSNR & SSIM & LPIPS & PSNR & SSIM & LPIPS & PSNR & SSIM & LPIPS \\
    \midrule
    NeILF++ \cite{zhang2023neilf++}
    & 31.482 & 0.9775 & 0.0591
    & 28.940 & 0.9498 & 0.0886
    & 29.657 & 0.9640 & 0.0676
    & 28.476 & 0.9336 & 0.0856
    & 24.729 & 0.8986 & 0.1202 \\
    WildLight \cite{cheng2023wildlight}
    & - & - & -
    & 29.913 & 0.9399 & 0.0787
    & 32.032 & 0.9669 & 0.0520
    & 28.219 & 0.9252 & 0.0981
    & 26.546 & 0.9078 & 0.1155 \\
    IRON \cite{iron-2022}
    & 31.845 & 0.9855 & 0.0320
    & 30.087 & 0.9550 & 0.0468
    & 31.713 & 0.9808 & 0.0366
    & 27.403 & 0.9602 & \textbf{0.0583}
    & 25.516 & 0.9257 & 0.0876 \\
    Ours
    & \textbf{35.164} & \textbf{0.9912} & \textbf{0.0273}
    & \textbf{32.979} & \textbf{0.9729} & \textbf{0.0340}
    & \textbf{33.921} & \textbf{0.9851} & \textbf{0.0327}
    & \textbf{29.209} & \textbf{0.9640} & 0.0591
    & \textbf{27.647} & \textbf{0.9391} & \textbf{0.0767} \\
    \bottomrule
\end{tabular}
}

%% file: tables/res_material_estimation.tex
\resizebox{0.85\textwidth}{!}{
\begin{tabular}{l||*{1}{c}|*{3}{c}|*{3}{c}|*{3}{c}}
    \toprule
    & \multicolumn{1}{c|}{Roughness}
    & \multicolumn{3}{c|}{Diffuse Albedo}
    & \multicolumn{3}{c|}{Specular Albedo}
    & \multicolumn{3}{c}{View Synthesis RGB}\\
    Method & MSE $\times 10^{-3}$ & PSNR & SSIM & LPIPS & PSNR & SSIM & LPIPS & PSNR & SSIM & LPIPS \\
    \midrule
    WildLight \cite{cheng2023wildlight} & 106.32
    & 25.631 & 0.9189 & 0.1186
    & 17.357 & 0.8353 & 0.2016
    & 29.167 & 0.9300 & 0.0929 \\
    IRON \cite{iron-2022} & 1.8402
    & 33.175 & 0.9730 & 0.0432
    & 25.809 & 0.8496 & 0.1645
    & 29.053 & 0.9604 & 0.0558
    \\
    Ours & \textbf{0.8808}
    & \textbf{35.777} & \textbf{0.9805} & \textbf{0.0331}
    & \textbf{29.716} & \textbf{0.9136} & \textbf{0.1065}
    & \textbf{31.891} & \textbf{0.9700} & \textbf{0.0488} \\
    \bottomrule
\end{tabular}
}

%% file: tables/res_collocated_synthetic_material.tex
\resizebox{0.85\textwidth}{!}{
\begin{tabular}{l||*{1}{c}|*{3}{c}|*{3}{c}|*{3}{c}}
    \toprule
    & \multicolumn{1}{c|}{Roughness}
    & \multicolumn{3}{c|}{Diffuse Albedo}
    & \multicolumn{3}{c|}{Specular Albedo}
    & \multicolumn{3}{c}{View Synthesis RGB} \\
    Method & MSE $\times 10^{-3}$ & PSNR & SSIM & LPIPS & PSNR & SSIM & LPIPS & PSNR & SSIM & LPIPS \\
    \midrule
    WildLight \cite{cheng2023wildlight} & 104.62
    & 27.449 & 0.7979 & 0.2320
    & 19.518 & 0.8420 & 0.1747
    & 30.520 & 0.8959 & 0.0943 \\
    IRON \cite{iron-2022} & 0.8264
    & 33.616 & 0.9793 & 0.0419
    & 31.231 & 0.9159 & 0.1084
    & 33.827 & 0.9743 & 0.0405 \\
    Ours w/o Inter-reflect. & 0.8112
    & 34.966 & 0.9807 & 0.0318
    & 32.500 & 0.9250 & 0.0951
    & 34.382 & 0.9753 & 0.0393  \\
    Ours w/o $\boldsymbol{f}_{\text{dino}}$ & 0.7638
    & 34.806 & 0.9812 & 0.0327
    & 32.526 & 0.9182 & 0.1005
    & 34.294 & 0.9755 & 0.0393  \\
    Ours & \textbf{0.6226}
    & \textbf{35.075} & \textbf{0.9817} & \textbf{0.0316}
    & \textbf{33.128} & \textbf{0.9400} & \textbf{0.0854}
    & \textbf{34.804} & \textbf{0.9762} & \textbf{0.0391}  \\
    \bottomrule
\end{tabular}
}

%% file: sec/6_conclusions.tex
\section{Conclusion}
\label{sec:Conclusion}
In this paper, we present an effective inverse rendering approach for reconstructing object shapes, materials, and lighting from photometric images.
Our method optimizes the light source position to account for self-shadows and employs an online strategy for modeling inter-reflections through a differentiable rendering layer.
Additionally, we incorporate the DINO regularization to help the decomposition of surface reflectance.
Extensive experiments on synthetic and real datasets demonstrate that our method can address misalignments between camera and light sources and surpass state-of-the-art methods in material decomposition.

\paragraph{Limitations} 
Our method overlooks the consistency of novel views captured by a moving flashlight during geometry initialization, and the BRDF model is tailored for solid reflective surfaces. Future work will address these limitations and extend our approach to more complex imaging scenarios, including underwater environments~\cite{zhang2023beyond,ju2024underwater}.


\paragraph{Acknowledgment} This work was supported in part by NSFC with Grant No.~62293482, the Basic Research Project No.~HZQB-KCZYZ-2021067 of Hetao Shenzhen-HK S\&T Cooperation Zone. It was also partially supported by NSFC with Grant No. 62202409, the Shenzhen Science and Technology Program with Grant No. RCBS20221008093241052, the Shenzhen Outstanding Talents Training Fund 202002, the Guangdong Research Projects No.~2017ZT07X152 and No.~2019CX01X104, the Guangdong Provincial Key Laboratory of Future Networks of Intelligence (Grant No.~2022B1212010001), and the Shenzhen Key Laboratory of Big Data and Artificial Intelligence (Grant No.~ZDSYS201707251409055).

%% file: sec/7_supp.tex
\clearpage
\setcounter{page}{1}
\setcounter{section}{0}
\maketitlesupplementary

\section{More Details for the Method}

\subsection{Network Architectures}

\paragraph{Neural SDF:}$f_{\Theta_g} (\point) = (s, \fgeo)$. 
We employ an 8-layer MLP featuring a hidden dimension of 256 and incorporate a skip connection at the fourth layer. The network input is the 3D coordinate $\point$ encoded with a frequency of 6, to output the SDF value and an implicit local geometric feature. Before optimization, we perform geometric initialization on the network, as described by \cite{Atzmon_2020_CVPR}.

\paragraph{Neural diffuse albedo:}$f_{\Theta_d}(\boldsymbol{x}, \boldsymbol{n}, \boldsymbol{n}, \boldsymbol{f}) = \rho_d$.
We use an 8-layer MLP featuring a hidden dimension of 256 and a skip connection at the fourth layer. The network inputs include the 3D coordinate $\point$ encoded with 10 frequencies, surface normal, and geometric features. It outputs the diffuse albedo for point $\point$.

\paragraph{Neural specular albedo:}$f_{\Theta_s}(\boldsymbol{x}, \boldsymbol{n}, \boldsymbol{f}) = \rho_s$.
We employ a 4-layer MLP with a width of 256. The input 3D coordinate $\point$ is encoded using 6 frequencies.

\paragraph{Neural roughness:}$f_{\Theta_r}(\boldsymbol{x}, \boldsymbol{n}, \fgeo) = \rho_r$.
We deploy a 4-layer MLP with a width of 256. The input 3D coordinate $\point$ is encoded using 6 frequencies.


\paragraph{Blending scalar:}$\gamma\left(\left\|\boldsymbol{x}-\boldsymbol{x}^{\prime}\right\|, \boldsymbol{w}_i \cdot \boldsymbol{n}, \rho_r\right) = \gamma$. We use a 4-layer MLP with a width of 128. The dot product of normal and view direction uses 6 frequencies. 

\paragraph{Neural DINO feature:}$f_{\Theta_\text{dino}}(\point) = \fdino$
We utilize a 4-layer MLP with a width of 256, where the input location $\point$ is encoded with 6 frequencies, and the output features a dimension of 384.

\subsection{Visibility Computation}

For joint optimization of object geometry and light position, we determine the visibility of a surface point $\point$ by uniformly sampling $N=128$ points $\{\point_i\}_{i=1}^{N}$ along the path from surface point $\point$ to the light source. We obtain the discrete opacity values $\{\alpha\}_{i=1}^{N}$ for these points using the unbiased SDF density conversion method introduced by NeuS \cite{wang2021neus}:
\begin{align}
\alpha_i=\max \left(\frac{\Phi_s\left(f\left(\mathbf{p}\left(t_i\right)\right)\right)-\Phi_s\left(f\left(\mathbf{p}\left(t_{i+1}\right)\right)\right)}{\Phi_s\left(f\left(\mathbf{p}\left(t_i\right)\right)\right)}, 0\right).
\end{align}

The light visibility of point $\boldsymbol{x}$ in the direction of incident light $\boldsymbol{w}_i$ is represented by the residual transmittance:
\begin{align}
f_v\left(\boldsymbol{w}_i ; \boldsymbol{x}\right)=1-\sum_{j=1}^N \alpha_j T_j ,
\end{align}
where $\alpha_j$ is the density value at point $\boldsymbol{x}_{j}$, and $T_j = \prod_{k=1}^{j-1}\left(1-\alpha_k\right)$ is the light transmittance at point $\boldsymbol{x}_{j}$ in the direction $\boldsymbol{w}_i$.

\subsection{Inter-reflection Computation}

\paragraph{Importance Sampling} To model the indirect illumination in scenes dynamically captured with directional lighting, we introduce an online computation approach that combines a differentiable layer and an importance sampling strategy. For a point $\point$ and view direction $\boldsymbol{w}_i$, we consider a single light bounce and employ a ray marching towards the reflective direction:
\begin{align}
\boldsymbol{w}_r = 2 \times \boldsymbol{n} - \boldsymbol{w}_i .
\end{align}
We then identify the secondary intersection point $\point'$. To determine if $\boldsymbol{x}'$ is occluded from the light source, we uniformly sample 20 points along the path between the light source and the intersection point. The light is considered occluded by another surface if any of the sampled points exhibit a negative SDF value.

\begin{figure*}[tb] \centering
    \includegraphics[width=0.9\linewidth]{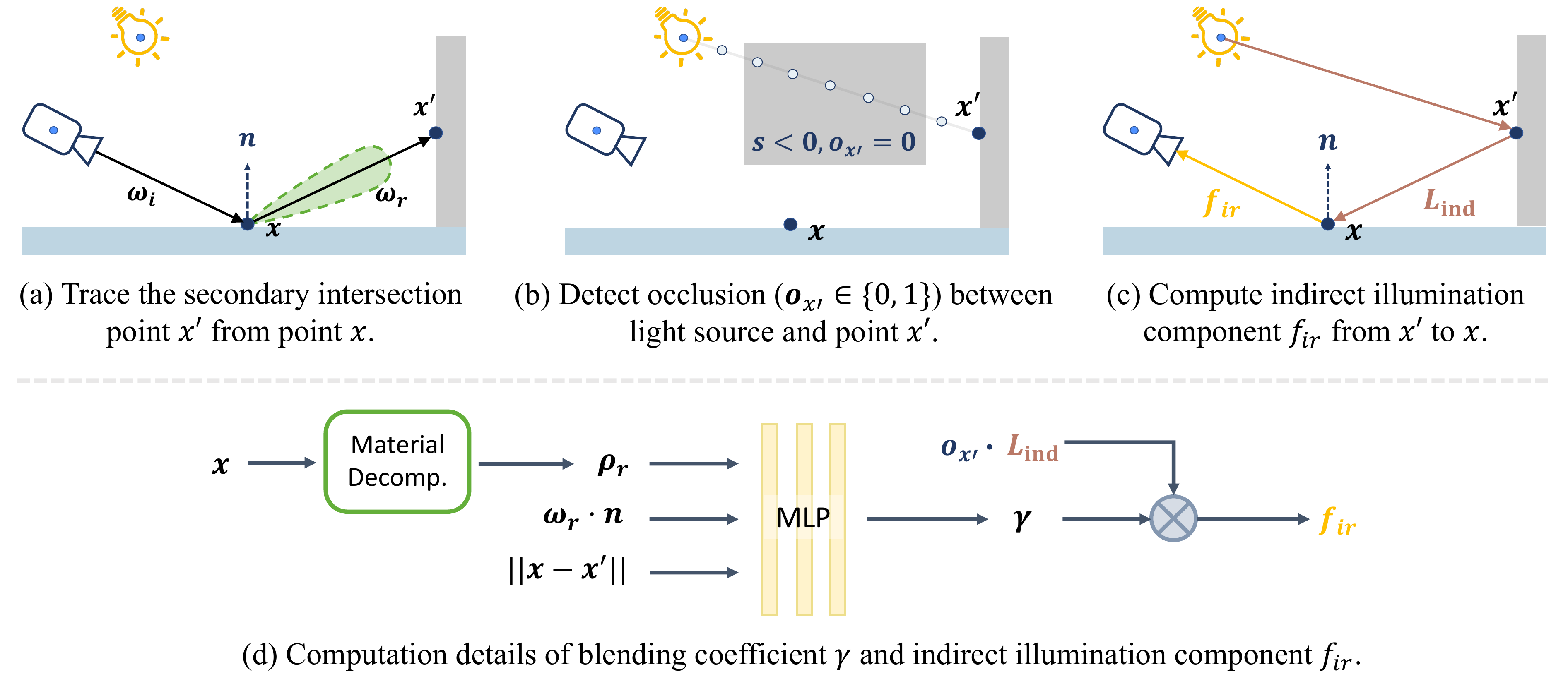}
    \caption{The illustration of inter-reflection modeling. We take into account rays that are physically rendered at secondary intersection points near the reflective direction of point $\point$, as these rays contribute significantly to the indirect illumination.} \label{fig:inter}
\end{figure*}

\Fref{fig:inter} illustrates the process of inter-reflection modeling. If the secondary intersection point $\point'$ is unobstructed, we compute the outgoing radiance at $\point'$ using the flashlight's incoming radiance. The outgoing result is then combined with the blending coefficient to represent the indirect illumination.

\paragraph{Gradient Backpropogation} Given that the blending coefficient is conditioned on the roughness property of the 3D point, there exists a correlation between the roughness property and the blending coefficient, introducing additional ambiguity in material estimation. 
In our experiments, we found that detaching the roughness of the secondary intersection point $\point'$ prior to its input into the blending coefficient network leads to a more precise material decomposition. Moreover, the process of gradient backpropagation starts from the image loss, through the residual component, and into the material networks of the secondary intersection point $\point'$, fostering the alignment of secondary point radiance with inter-reflection cues. 
In our experiments, we discovered that disabling the optimization of local geometry at the secondary point reduces the complexity of the optimization process, leading to improved geometric reconstruction, especially in concave areas.

\subsection{BRDF Renderer}

Our BRDF implementation closely adheres to the Mitsuba roughplastic BRDF model \cite{jakob2010mitsuba}, with the distribution parameter specifically set to `ggx'. For simplicity, we refer to our configuration as the roughplastic model. Default values are maintained for the internal and external Indices of Refraction and the nonlinear parameter.

Previous methods (IRON~\cite{iron-2022} and WildLight~\cite{cheng2023wildlight}) employing the renderer relied on an oversimplified BRDF model within an idealized setting where the camera and flashlight are collocated, neglecting the deviations between the camera and light source present in our capture setup. To address this limitation, we have enhanced the simplified roughplastic model to accommodate a broader range of scenarios, allowing for variations in both incident and outgoing light directions.

\subsection{Training Details}




The training process requires approximately 9 hours on a single RTX3090 GPU with 24GB of memory.
We start by training NeuS over 100,000 iterations to initialize the geometry and diffuse albedo networks. For each training iteration, we utilize 512 randomly sampled pixels, employing an $\ell_1$ loss along with an eikonal regularization loss.
Prior to the rendering phase, we derive the feature maps of images by the pre-trained ViT-S/8 model \cite{caron2021emerging} and executed 10,000 iterations with $\lambda_4$ set to $1.0$ to extract the DINO feature from 2D feature maps to 3D surfaces.
During the physics-based surface rendering stage with a total of 50,000 iterations, we fixed the geometry and lighting to warm up the BRDFs network for 2,000 iterations to stabilize the process, and subsequently, we carried out a joint optimization of the lighting, geometry, and BRDFs. The training of the blending coefficient network started at the 10,000th iteration. We set the size of rendered image patch as $128\times 128$ and loss weights to $\lambda_1 = 10^{-4}$, $\lambda_2 = 0.1$, $\lambda_3 = 10^{-5}$ and $\lambda_4 = 10^{-5}$. All networks are optimized by corresponding Adam optimizers with learning rate $10^{-4}$.

\section{More Details for the Dataset}

\paragraph{DRV Dataset}
We acquired the DRV dataset \cite{bi2020deep} from the authors, comprising five scenes: \emph{Dragon}, \emph{Girl}, \emph{Pony}, \emph{Tree}, and \emph{Cartoon}. Each scene has approximately 400 images, split between training and test sets. The dataset captures images in a darkroom, utilizing a nearly collocated camera-light setup.

\paragraph{Luan Dataset}
The Luan dataset \cite{luan2021unified} was captured using a casual smartphone. We noticed that the images exhibit significant noise and motion blur, together with varying exposure times and white balance settings during capture. This inconsistency introduces challenges in maintaining multi-view consistency. We evaluated the scene \emph{Xmen}, which includes 136 images, to compare novel view rendering and material decomposition against the IRON method.

\begin{figure*}[t] \begin{center}
    \includegraphics[width=\linewidth]{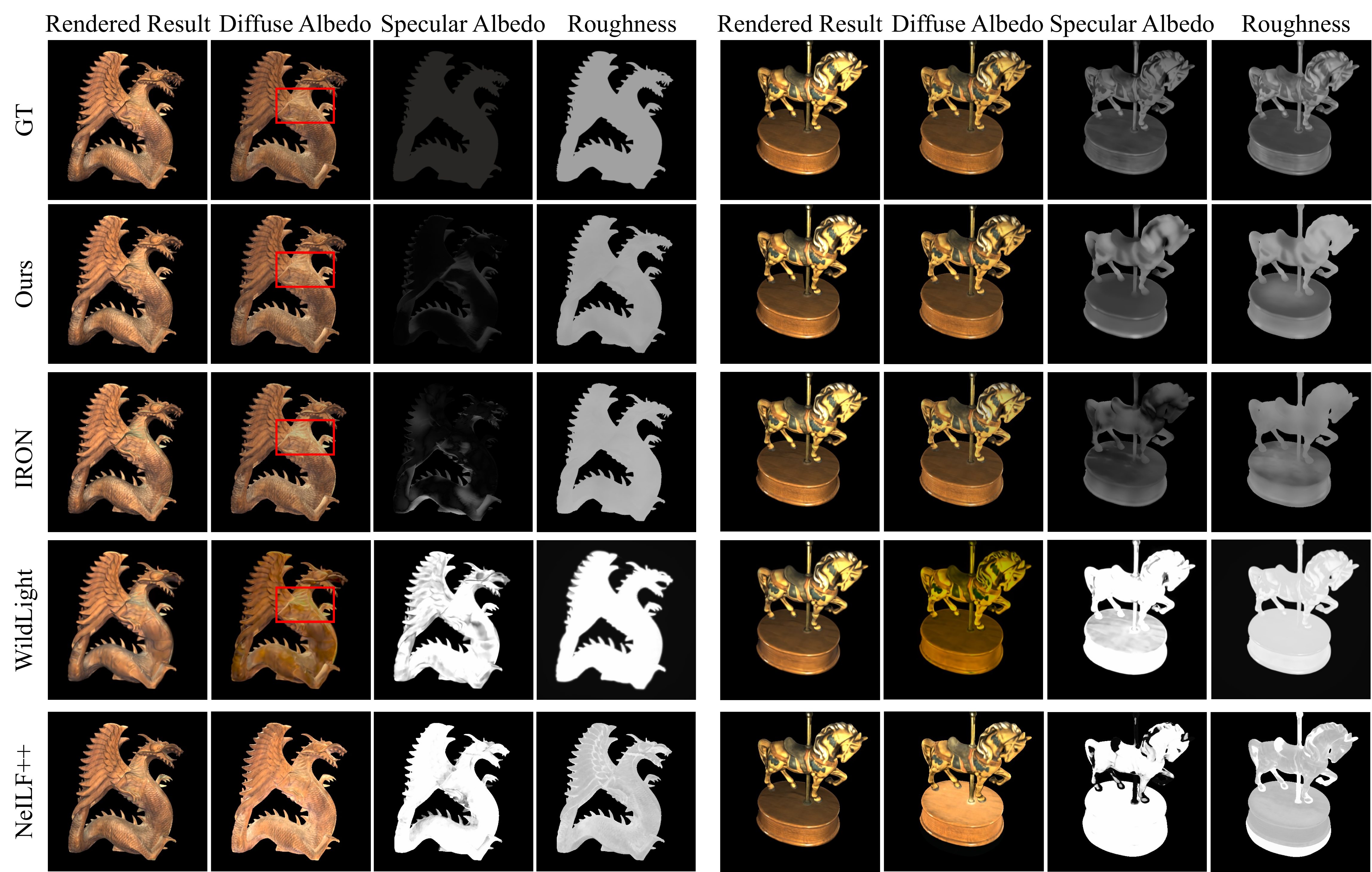}
    \caption{Qualitative comparisons with state-of-art methods on the synthetic dataset (\emph{dragon} and \emph{horse}). The materials of NeILF++\cite{zhang2023neilf++} are \emph{Base Color}, \emph{Metallic}, \emph{Roughness} defined by simplified Disney principled BRDF.}
    \label{fig:supp_synthetic}
    \vspace{-1.5em}
\end{center} \end{figure*}

\paragraph{Self-captured Dataset} For capturing real-world images, we employed an iPhone 15 to shoot in RAW format, ensuring a linear camera response. Across all photos, we maintained consistent settings for the camera's exposure time, focus, and white balance. Specifically, the ISO value and shutter speed (exposure time) were fixed at 100 and 1/250s, respectively, with the white balance adjusted to 3,800 Kelvin degrees. Our collection encompasses 5 scenes: \emph{Toy}, \emph{fruit}, \emph{Panda}, \emph{Assassin}, and \emph{Bear}, with the number of images per scene varying from 120 to 400. Camera poses were derived using COLMAP \cite{schonberger2016_sfm_cvpr16}, and objects were scaled to fit within a unit sphere based on the reconstructed point cloud. 
The photography sessions took place in a darkroom, positioning the camera 0.15 to 0.3 meters away from each object. To achieve comprehensive coverage, we systematically moved the camera in a spiral pattern around the subjects. The separation between the camera lens and the flashlight on the iPhone, roughly 0.015m, results in an approximate 3-degree variation between viewing and lighting angles at a standard distance of 0.25m from the object.



\section{More Results and Comparisons}


We primarily compare our results with those from IRON \cite{iron-2022} and WildLight \cite{cheng2023wildlight}. Notably, WildLight was unable to reconstruct the synthetic data for \emph{duck}, and as such, its results are not presented in the table within the main paper.

\subsection{Results on Synthetic Data}

\begin{table*}[t] \begin{center}
    \input{tables/supp-res_synthetic}
    \caption{Complete results on the synthetic dataset.}
    \label{tab:supp-res_synthetic}
    \input{tables/supp-res_real}
    \caption{Quantitative comparison of novel view rendering on DRV dataset.}
    \label{tab:supp-res_real}
    \vspace{-2em}
\end{center} \end{table*}

\begin{figure}
    \centering
    \vspace{-1em}
    \includegraphics[width=\linewidth]{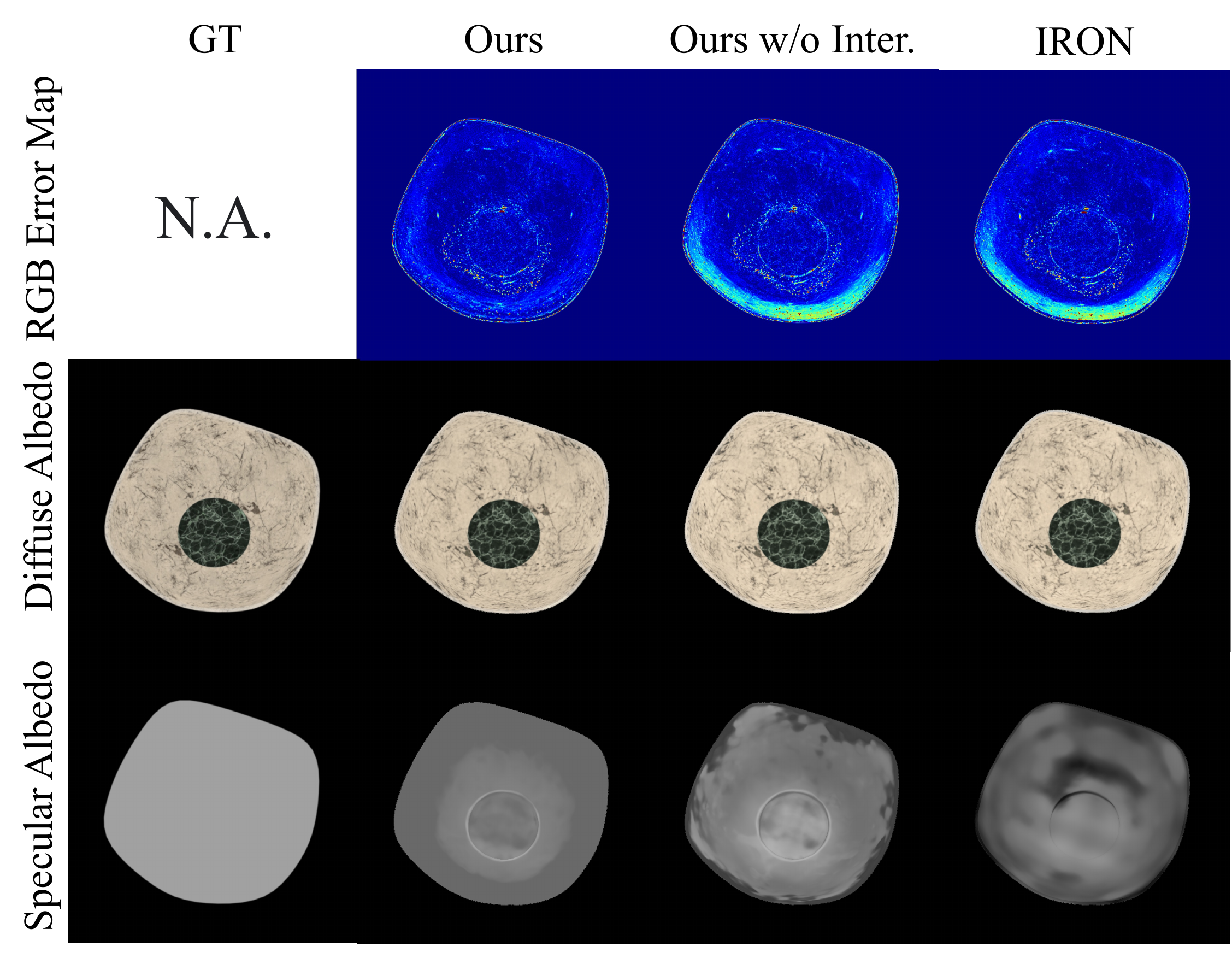}
    \caption{Ablation study on inter-reflection in \emph{marble bowl}.}
    \vspace{-2em}
    \label{fig:abla-inter2}
\end{figure}

In \Tref{tab:supp-res_synthetic}, we offer comprehensive results for each synthetic scene captured under casual conditions. Additionally, we provide a qualitative comparison of novel view rendering and material decomposition between our method and earlier methods, as illustrated in \fref{fig:supp_synthetic}. For the \emph{dragon} scene, our method produces a diffuse albedo with less indirect illumination incorporated into the materials. In the \emph{horse} scene, our material decomposition results demonstrate a reduced influence of self-shadows, showing a closer alignment with the ground truth than those obtained with IRON. Even in extremely concave regions, our method is more robust than the previous method, as shown in \fref{fig:abla-inter2}.



\subsection{Results on Real Data}

\begin{figure*}
    \includegraphics[width=0.96\linewidth]{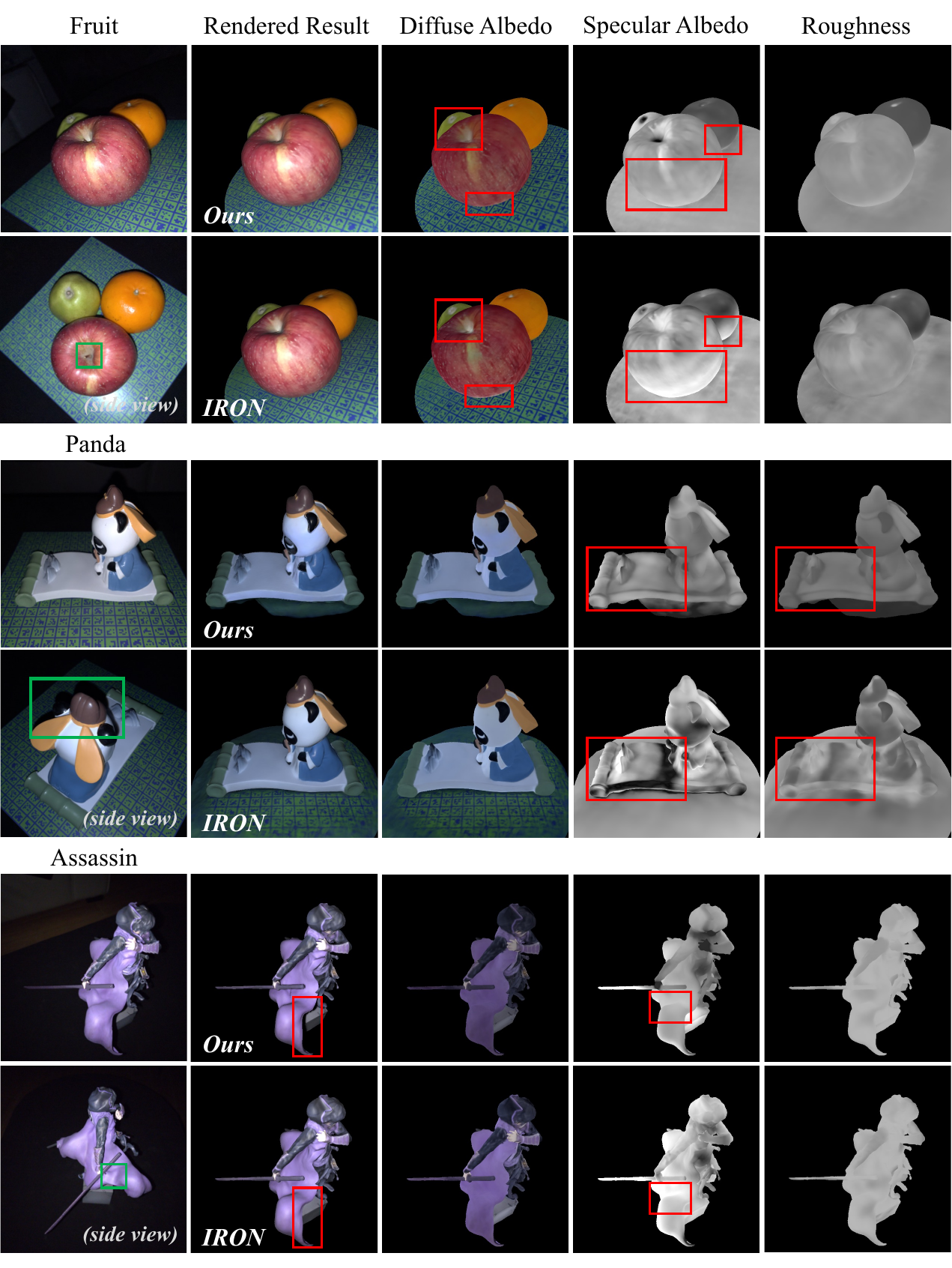}
    \vspace{-2.5em}
    \caption{More visual results of material decomposition on our dataset. }
    \label{fig:supp_res_real_ours}
\end{figure*}

In \fref{fig:supp_res_real_ours}, we present our dataset's novel view rendering and material decomposition outcomes. The IRON method often incorporates indirect illumination into the diffuse albedo, particularly in concave regions, as observed in the \emph{Fruit} scene. Additionally, specular albedoes produced by the IRON method are adversely affected by self-shadows and inter-reflections, as highlighted in specific boxes.

\begin{figure*}
    \includegraphics[width=0.96\linewidth]{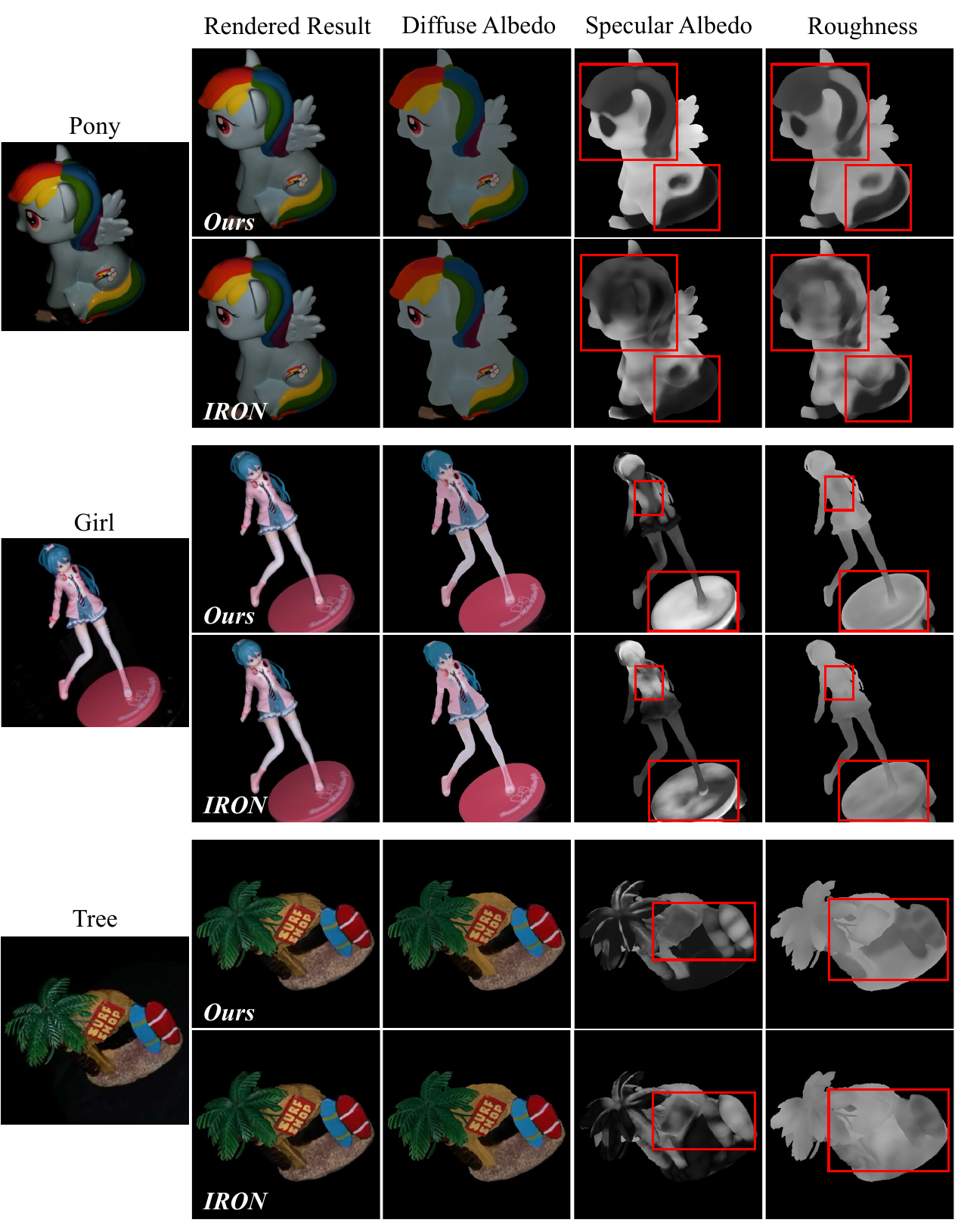}
    \caption{
    Visual results of material decomposition on DRV dataset. }
    \label{fig:supp_res_real_drv1}
\end{figure*}

\begin{figure*}
    \includegraphics[width=\linewidth]{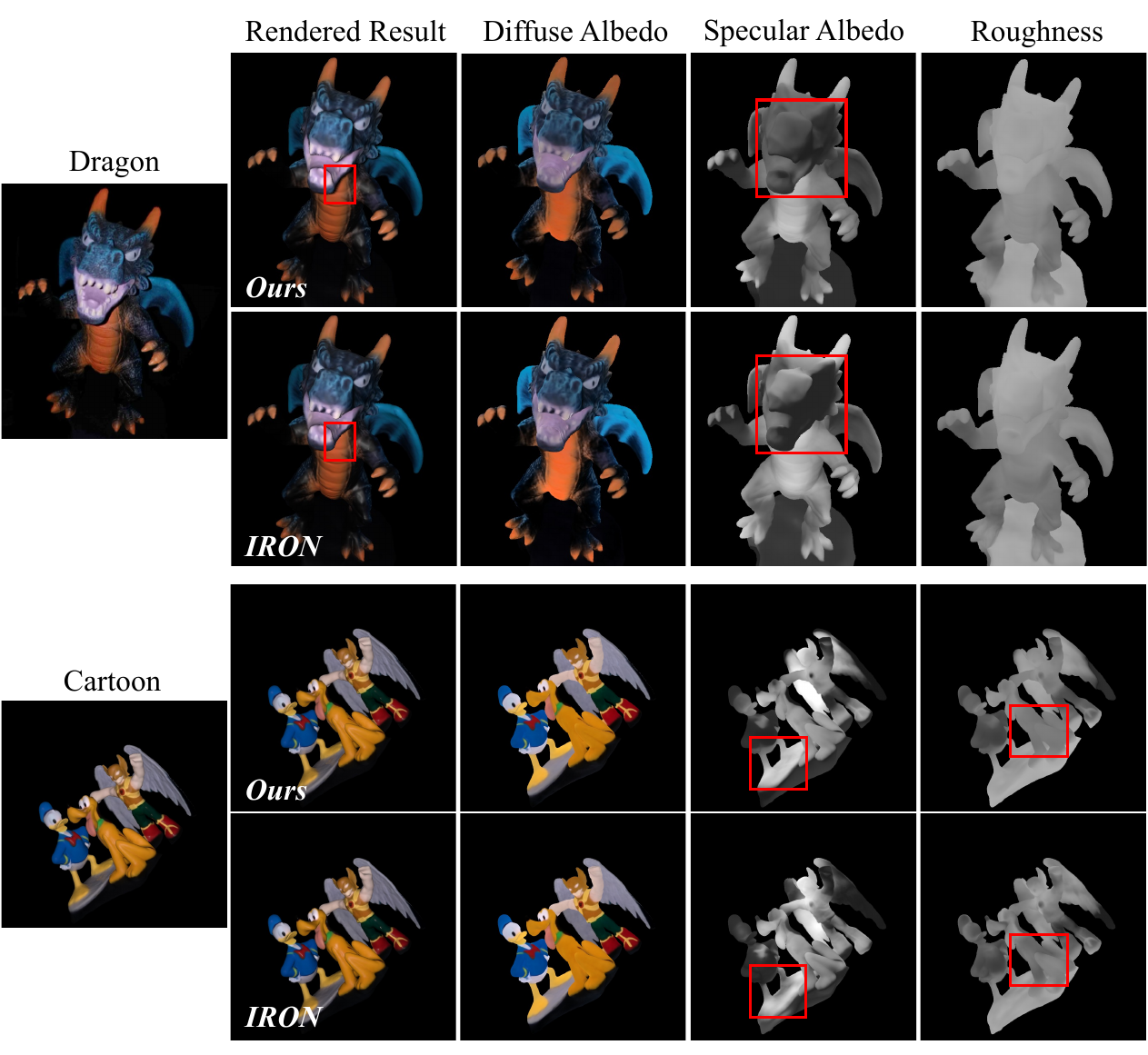}
    \caption{
    Visual results of material decomposition on DRV dataset (continued). }
    \label{fig:supp_res_real_drv2}
\end{figure*}

In \Tref{tab:supp-res_real}, we provide a quantitative comparison that underscores the enhanced performance of our method compared to IRON in terms of novel view rendering within the DRV real dataset. \Fref{fig:supp_res_real_drv1} and \fref{fig:supp_res_real_drv2} complement this with side-by-side qualitative comparisons of our method against IRON regarding material decomposition.
Leveraging DINO regularization for surface decomposition, which effectively clusters similar materials, our approach produces more accurate results for material decomposition, especially in scenarios with a skewed view distribution. We observe that IRON's evaluation metrics for the \emph{Dragon} scene slightly exceed those of our method, this disparity is primarily due to its collocated camera-lighting setup, which inherently minimizes the occurrence of self-shadows within the scene.

\subsection{Failure Case}
\begin{figure}
    \includegraphics[width=\linewidth]{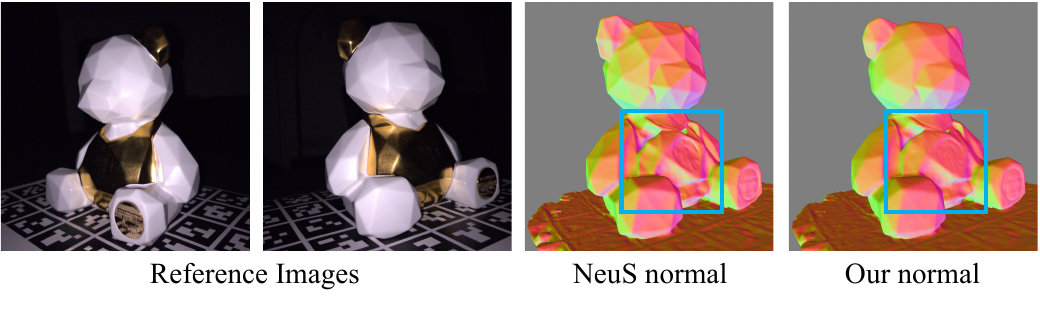}
    \caption{A failure case on \emph{bear} with reflective surfaces.}
    \label{fig:supp_fail}
    \vspace{-1em}
\end{figure}
Like many neural surface reconstruction methods, both COLMAP and NeuS presuppose Lambertian observation to guarantee multi-view consistency. Following the same approach as IRON, our method primarily depends on NeuS for geometry initialization but struggles to reconstruct objects with reflective surfaces, as depicted in \fref{fig:supp_fail}.  The surfaces reconstructed by NeuS and our method exhibit holes within reflective regions.

\section{Video Demos}

In the video, we present more comprehensive results to demonstrate the effectiveness of our design, along with additional comparison cases between our method and other inverse rendering methods. Furthermore, we render the reconstructed 3D assets using a traditional graphical pipeline to illustrate their practical applications.





%% file: tables/supp-res_synthetic.tex

\resizebox{\linewidth}{!}{
\begin{tabular}{@{} c c c c c c ccc c ccc c ccc@{}}
    \toprule
    \multirow{2}{*}{Scene} & &
    \multirow{2}{*}{Method} & & 
    \multicolumn{1}{c}{Roughness} & & \multicolumn{3}{c}{Diffuse Albedo} & & \multicolumn{3}{c}{Specular Albedo} & & \multicolumn{3}{c}{Novel View Synthesis}  \\ \cline{5-5} \cline{5-9} \cline{11-13} \cline{15-17}
    & & & & MSE $\times 10^{-3}$ $\downarrow$ & & PSNR $\uparrow$ & SSIM $\uparrow$ & LPIPS $\downarrow$ & & PSNR $\uparrow$ & SSIM $\uparrow$ & LPIPS $\downarrow$ & & PSNR $\uparrow$ & SSIM $\uparrow$ & LPIPS $\downarrow$ \\ \hline

    \multirow{3}{*}{\emph{duck}} & & WildLight & & - & & - & - & - & & - & - & - & & - & - & - \\
    
    & & IRON & & 2.640 & & 24.693 & 0.9631 & 0.0483 & & 18.669 & 0.9017 & 0.1075 & & 31.845 & 0.9855 & 0.0320 \\ 
                
    & & Ours & & \textbf{1.059} & & \textbf{34.871} & \textbf{0.9852} & \textbf{0.0355} & & \textbf{23.402} & \textbf{0.9474} & \textbf{0.0730} & & \textbf{35.164} & \textbf{0.9912} & \textbf{0.0273} \\ \hline

    \multirow{3}{*}{\emph{maneki}} & & WildLight & & 93.59 & & 18.151 & 0.7351 & 0.4472 & & 12.413 & 0.8114 & 0.2497 & & 29.913 & 0.9400 & 0.0787 \\
    
    & & IRON & & 1.455 & & 35.367 & 0.9805 & 0.0238 & & 18.967 & 0.8369 & 0.1732 & & 30.087 & 0.9550 & 0.0468 \\ 
    
    & & Ours & & \textbf{0.467} & & \textbf{36.098} & \textbf{0.9880} & \textbf{0.0184} & & \textbf{22.245} & \textbf{0.9371} & \textbf{0.0726} & & \textbf{32.979} & \textbf{0.9729} & \textbf{0.0340} \\ \hline

    \multirow{3}{*}{\emph{horse}} & & WildLight & & 40.23 & & 24.625 & 0.9507 & 0.1032 & & 16.997 & 0.8401 & 0.2056 & &  32.032 & 0.9669 & 0.0520 \\
    
    & & IRON & & 2.198 & & 31.903 & 0.9826 & 0.0363 & & 29.323 & 0.8701 & 0.1275 & & 31.713 & 0.9808 & 0.0366 \\ 
    
    & & Ours & & \textbf{1.509} & & \textbf{33.573} & \textbf{0.9880} & \textbf{0.0194} & & \textbf{33.071} & \textbf{0.9259} & \textbf{0.0917} & & \textbf{34.206} & \textbf{0.9831} & \textbf{0.0321} \\ \hline
    
    \multirow{3}{*}{\emph{marble bowl}} & & WildLight & & 165.2 & & 22.613 & 0.8862 & 0.1379 & & 15.600 & 0.8261 & 0.2135 & & 28.219 & 0.9252 & 0.0981 \\
    
    & & IRON & & 0.321 & & 29.258 & 0.9623 & 0.0518 & & 35.035 & 0.8947 & 0.1553 & & 27.403 & 0.9602 & \textbf{0.0583} \\ 
    
    & & Ours & & \textbf{0.172} & & \textbf{29.881} & \textbf{0.9647} & \textbf{0.0493} & & \textbf{39.972} & \textbf{0.9729} & \textbf{0.0660} & & \textbf{29.209} & \textbf{0.9640} & 0.0591 \\ \hline

    \multirow{3}{*}{\emph{dragon}} & & WildLight & & 120.8 & & 33.679 & 0.9208 & 0.1108 & & 14.432 & 0.7840 & 0.2453 & & 26.546 & 0.9078 & 0.1155 \\
    
    & & IRON & & 2.815 & & 36.470 & 0.9675 & 0.0575 & & 30.902 & 0.7610 & 0.2295 & & 25.516 & 0.9257 & 0.0876 \\ 
    
    & & Ours & & \textbf{0.923} & & \textbf{38.720} & \textbf{0.9735} & \textbf{0.0390} & & \textbf{34.894} & \textbf{0.8152} & \textbf{0.1772} & & \textbf{27.870} & \textbf{0.9406} & \textbf{0.0766} \\ \hline

    \multirow{3}{*}{\emph{armchair}} & & WildLight & & 117.9 & & 30.116 & 0.9242 & 0.1282 & & 15.748 & 0.8260 & 0.2136 & & 29.126 & 0.9100 & 0.1200 \\
    
    & & IRON & & 1.612 & & 41.361 & 0.9818 & 0.0416 & & 21.960 & 0.8333 & 0.1938 & & 27.752 & 0.9555 & 0.0734 \\ 
    
    & & Ours & & \textbf{1.155} & & \textbf{41.518} & \textbf{0.9833} & \textbf{0.0370} & & \textbf{25.442} & \textbf{0.8959} & \textbf{0.1438} & & \textbf{31.916} & \textbf{0.9655} & \textbf{0.0642} \\
    \bottomrule
\end{tabular}
}

%% file: tables/supp-res_real.tex
\resizebox{0.8\textwidth}{!}{
\begin{tabular}{l||*{2}{c}|*{2}{c}|*{2}{c}|*{2}{c}|*{2}{c}|*{2}{c}}
    \toprule
    & \multicolumn{2}{c|}{\emph{Pony}} 
    & \multicolumn{2}{c|}{\emph{Girl}}
    & \multicolumn{2}{c|}{\emph{Tree}}
    & \multicolumn{2}{c|}{\emph{Dragon}}
    & \multicolumn{2}{c|}{\emph{Cartoon}}
    & \multicolumn{2}{c}{Average}\\
    Method & PSNR & SSIM & PSNR & SSIM & PSNR & SSIM & PSNR & SSIM & PSNR & SSIM & PSNR & SSIM \\
    \midrule
    IRON \cite{iron-2022}
    & 29.269 & 0.9150 
    & 27.136 & 0.9326 
    & 31.641 & \textbf{0.9464} 
    & \textbf{32.421} & \textbf{0.9317} 
    & 30.773 & 0.9587 
    & 30.248 & 0.9369 \\
    Ours
    & \textbf{30.092} & \textbf{0.9414}
    & \textbf{27.589} & \textbf{0.9365}
    & \textbf{31.765} & \textbf{0.9464}
    & 32.251 & 0.9306
    & \textbf{30.975} & \textbf{0.9589}
    & \textbf{30.534} & \textbf{0.9428}
    \\
    \bottomrule
\end{tabular}
}